\documentclass[10pt,journal,compsoc]{IEEEtran}

\usepackage{times}
\usepackage{epsfig}
\usepackage{graphicx}
\usepackage{amsmath}
\usepackage{amssymb}
\usepackage{bbding}
\usepackage{pifont}
\usepackage{wasysym}
\usepackage{multirow}
\usepackage{color, xcolor}
\usepackage{rotating}
\usepackage{array}
\usepackage{subcaption}
\usepackage{booktabs}
\usepackage{bbm}
\usepackage{mdwlist}
\usepackage{float}
\usepackage{url}
\usepackage{cite}
\usepackage{amsfonts}
\usepackage{threeparttable}
\usepackage{enumerate}
\usepackage[noend]{algpseudocode}
\usepackage{algorithmicx,algorithm}
\usepackage{ragged2e}
\usepackage{pifont}         
\newcommand{\cmark}{\ding{51}}
\newcommand{\xmark}{\ding{55}}

\usepackage{caption}
\newlength\savewidth\newcommand\shline{\noalign{\global\savewidth\arrayrulewidth
  \global\arrayrulewidth 1pt}\hline\noalign{\global\arrayrulewidth\savewidth}}

\usepackage[pagebackref=true,breaklinks=true,colorlinks,bookmarks=false]{hyperref}
\ifCLASSINFOpdf
\else
\fi
\begin{document}




\renewcommand{\baselinestretch}{0.998}
\renewcommand{\justify}{\leftskip=0pt \rightskip=0pt plus 0cm}



%
\title{Hierarchical Prompts  for Rehearsal-free Continual Learning}
%
%
%
%

\author{Yukun Zuo,
        Hantao~Yao,~\IEEEmembership{Member,~IEEE},
        Lu~Yu,~\IEEEmembership{Member,~IEEE},
        Liansheng~Zhuang,~\IEEEmembership{Member,~IEEE},
        and~Changsheng~Xu,~\IEEEmembership{Fellow,~IEEE}

\thanks{Yukun Zuo and Liansheng Zhuang are with the School of Information Science and Technology, University of Science and Technology of China, Hefei, 230026, China, Email:zykpy@mail.ustc.edu.cn, lszhuang@ustc.edu.cn}
\thanks{Hantao Yao  is with National Laboratory of Pattern Recognition, Institute of Automation, Chinese Academy of Sciences, Beijing, 100190, China, Email:hantao.yao@nlpr.ia.ac.cn}
\thanks{Lu Yu is with the School of Computer Science and Engineering,
Tianjin University of Technology, Tianjin 300384, China, Email:
luyu@email.tjut.edu.cn}
\thanks{Changsheng Xu is with National Laboratory of Pattern Recognition, Institute of Automation, Chinese Academy of Sciences, Beijing, 100190, China, and also with the School of Artificial Intelligence, University of Chinese Academy of Sciences, Beijing 100049, China, Email: csxu@nlpr.ia.ac.cn}
}

%
%

\markboth{Journal of \LaTeX\ Class Files,~Vol.~14, No.~8, August~2015}%
{Shell \MakeLowercase{\textit{et al.}}: Bare Advanced Demo of IEEEtran.cls for IEEE Computer Society Journals}
%



\IEEEtitleabstractindextext{%
\begin{abstract}
  \justify{
    Continual learning endeavors to equip the model with the capability to integrate current task knowledge while mitigating the forgetting of past task knowledge. 
    Inspired by  prompt tuning, prompt-based  methods  maintain a frozen backbone and train with slight learnable prompts to minimize the catastrophic forgetting that arises due to updating a large number of backbone parameters. 
    Nonetheless, these learnable prompts tend to concentrate on the discriminatory knowledge of the current task while ignoring past task knowledge, leading to that learnable prompts still suffering from  catastrophic forgetting.   
    This paper introduces a novel rehearsal-free paradigm for continual learning termed Hierarchical Prompts (H-Prompts), comprising three categories of prompts -- class prompt, task prompt, and general prompt. 
    To effectively depict the knowledge of past classes, class prompt leverages  Bayesian Distribution Alignment to model the distribution of classes in each task.
    To reduce the forgetting of past task knowledge, task prompt employs Cross-task Knowledge Excavation to amalgamate the knowledge encapsulated in the learned class prompts of past tasks and current task knowledge.
    Furthermore, general prompt utilizes Generalized Knowledge Exploration to deduce highly generalized knowledge in a self-supervised manner.
    Evaluations on two benchmarks substantiate the efficacy of the proposed H-Prompts, exemplified by an average accuracy of 87.8\% in Split CIFAR-100 and 70.6\% in Split ImageNet-R.}
\end{abstract}

\begin{IEEEkeywords}
  Continual learning; Rehearsal-free; Hierarchical Prompts.
\end{IEEEkeywords}

}

\maketitle

\IEEEdisplaynontitleabstractindextext

%
\IEEEpeerreviewmaketitle

\ifCLASSOPTIONcompsoc
\IEEEraisesectionheading{\section{Introduction}\label{sec:introduction}}
\else
\section{Introduction}
\fi

%
%
%
%


Deep learning has accomplished exceptional performance in various areas~\cite{a3,a4,a5}, but it falls into the paradigm of training a model on independent and identically distributed (i.i.d.) data. 
The inability to handle non-stationary data distribution under the current paradigm necessitates the retraining of a model whenever new tasks or categories continuously emerge, leading to  excessive training costs.
However, fine-tuning the trained model of past tasks on  current task makes the model encounter catastrophic forgetting~\cite{a2} -- a heavily deteriorated performance on previously encountered data.
To tackle this problem, continual learning~\cite{a6,a7,9349197,boschini2022class,10.1109/TPAMI.2017.2773081} strives to train a model on sequential tasks in a new paradigm, \emph{i.e.}, acquiring the knowledge of continuously incoming tasks while preserving the knowledge of previously learned tasks.

\begin{figure}
  \centering
  \begin{subfigure}{0.49\linewidth}
    \includegraphics[width=1\linewidth]{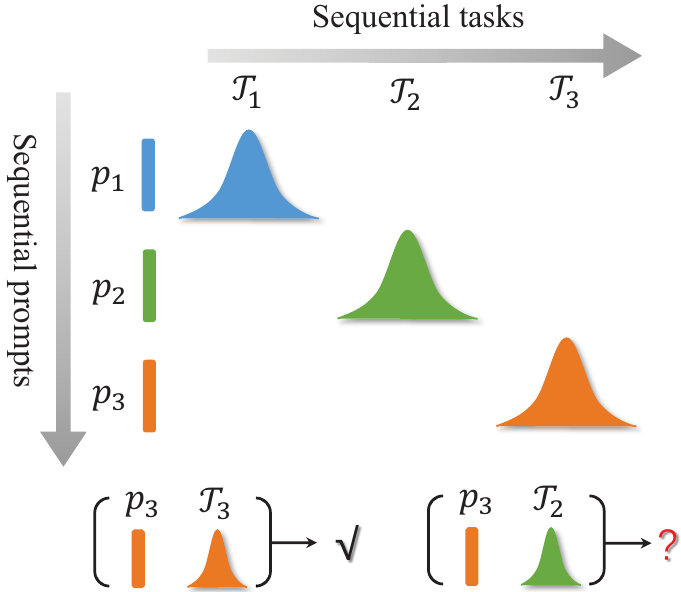}
    \caption{Previous methods}
  \end{subfigure}
  \begin{subfigure}{0.49\linewidth}
  \includegraphics[width=1\linewidth]{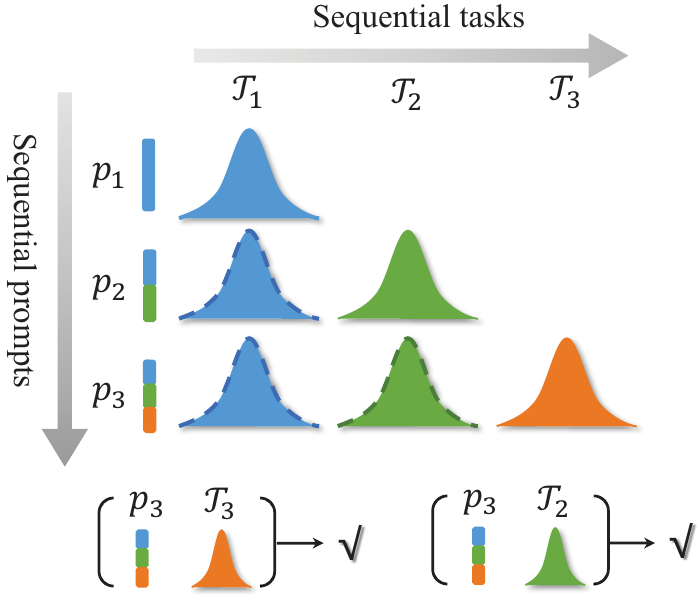}
  \caption{H-Prompts}
  \end{subfigure}
  \caption{(a) Previous prompt-based methods train prompt to focus on the knowledge of  current task, while ignoring the  knowledge of past tasks.
  (b) H-Prompts  stores and integrates past task knowledge with current task knowledge during prompt tuning.}
  \label{fig1}
  \end{figure}

A plethora of methods have been proposed to surmount catastrophic forgetting.
Regularization-based methods~\cite{a8,a9,a10,a11} quantify the pivotal parameters of past tasks and restrict their changes in the current task. 
Exemplar-based methods~\cite{a12,a13,a14,a15} store a portion of  historical exemplars to replay the historical knowledge.
Dynamic architecture methods~\cite{a16,a17,a18,a19}  expand new trainable modules for new tasks continually to  maintain the knowledge of past tasks.
Despite these efforts, these methods still experience substantial catastrophic forgetting of past tasks or burgeoning computational costs due to the updating of numerous model parameters. 
Recently, prompt-based methods~\cite{a20,a21} have garnered extensive attention because of the demand for slight learnable $prompt$~\cite{a22} parameters over the frozen backbone.
They leverage these learnable prompts to encode the task knowledge during training and select pertinent prompts during inference. 
However, the learned prompts tend to solely focus on acquiring discriminatory knowledge of the current task, neglecting past task knowledge, leading to failure in classifying historical images shown in Figure~\ref{fig1}(a).

To address the above problem, an intuitive idea is to explicitly consider historical knowledge while learning the learnable prompts on current task.
Specifically, we utilize the class-related statistic information, \emph{i.e.,}  learnable mean vector and diagonal covariance matrix of each class in sequence embedding level defined as \emph{class prompt}, to embed the historical knowledge of each class.
Furthermore, we replay the historical class prompts for making a learnable \emph{task prompt} of current task contain the knowledge of past tasks and current task simultaneously, shown in Figure~\ref{fig1}(b).
Additionally, leveraging highly generalized knowledge can also mitigate catastrophic forgetting in continual learning~\cite{a71,a51}.
Since self-supervised learning~\cite{a28,a29,a30} has been proven to obtain a model with high discrimination and generalization ability, we apply self-supervised learning to infer \emph{general prompt} to capture generalized knowledge.

In this work, we  propose a novel Hierarchical Prompts (H-Prompts) approach,  consisting of class prompt, task prompt, and general prompt.
Task prompt and general prompt are extended with input to instruct frozen backbone with generalized  knowledge and past task knowledge as well as current task knowledge,  respectively.
Moreover, class prompt replacing the position of input is extended with fixed task prompt and general prompt to preserve past task knowledge.
Therefore, the key aspects of H-Prompts  are to preserve the knowledge of past tasks for class prompt, capture past and current task knowledge for task prompt,  and learn generalized  knowledge for general prompt.
Specifically, to preserve the knowledge of past tasks, we propose Bayesian Distribution Alignment for class prompt to model the distribution of classes in each task with  Bayesian Neural Networks (BNNs) in an adversarial manner.
With the obtained class prompt, we present Cross-task Knowledge Excavation to integrate the past knowledge replayed by class prompt with current task knowledge into task prompt.
Furthermore, targeting to learn  generalized  knowledge, we perform Generalized Knowledge Exploration to conduct self-supervised learning for general prompt.

The contribution of this paper are summarized as follows: 
1) We propose a novel rehearsal-free   Hierarchical Prompts (H-Prompts) paradigm consisting of  class prompt, task prompt, and general prompt to overcome the catastrophic forgetting of prompts.  
2) We present Bayesian Distribution Alignment, Cross-task  Knowledge Excavation, and Generalized Knowledge Exploration to preserve past task knowledge for class prompt, learn past and current task knowledge for task prompt, and obtain  generalized  knowledge for general prompt, respectively. 
3) The proposed H-Prompts  achieves the best performance  on two standard class incremental learning benchmarks, \emph{e.g.}, obtaining the average accuracy of 87.8\% and  70.6\% in Split CIFAR-100 and  Split ImageNet-R,  respectively.

\section{Related Work}

\subsection{Prompt Learning}
Pormpt learning~\cite{a37,a44}   is initially introduced to prepend a fixed function to input text of pre-trained language model~\cite{a41}  in natural language processing (NLP).
This approach enables the pre-trained language model to receive additional instructions for adaptation in downstream tasks. 
Nevertheless, designing an appropriate prompting function is challenging and often necessitates expert knowledge. As a result, numerous subsequent studies~\cite{a45, a46} have explored strategies for creating more effective prompting texts.
Prompt tuning~\cite{a47} makes task-specific continual vectors as prompt, and updates the prompt with gradient descent to  encode task-specific knowledge.
Prompt learning  has also achieved promising performance in Computer Vision (CV)~\cite{a48,a49,a50}, demonstrating its strong generalization abilities in few-shot learning and zero-shot learning~\cite{a51,a52,a53}. 
Recently, L2P~\cite{a20} and Dualprompt~\cite{a21}  apply  prompt learning  into  continual learning for encoding the knowledge of each task, and achieve perferable performance in class incremental learning benchmarks.
However,  they still suffer from the problem of  catastrophic forgetting in  prompts.
For solving the above problem,  we propose  H-Prompts to  overcome the catastrophic forgetting of prompts via modeling the distribution of past classes,  and learn more generalized features via self-supervised learning.

\subsection{Continual Learning}

A variety of methods have been proposed to address the catastrophic forgetting of continual learning, which can be divided into non-prompt methods~\cite{a8,a11,a12,a25,a19,a71} and prompt-based methods~\cite{a20,a21}

\textbf{Non-prompt Methods}.
We highlight three primary categories of non-prompt methods: regularization-based methods, exemplar-based methods, and dynamic architecture methods.
Regularization-based methods~\cite{a8,a9,a10,a11} strive to identify critical parameters of past tasks and  constrain their update in the learning of current tasks.
However, these methods face challenges in efficiently assessing the importance of each parameter, which often results in suboptimal performance.
Exemplar-based methods~\cite{a12,a15,a35,a24,a25,Petit_2023_WACV,zhou2023model,sun2023regularizing} mitigate the degree of forgetting by replaying a small set of historical images  saved in memory~\cite{a12,a34,a14,a15,a35}, or replaying generated images  from Generative Adversarial Networks~\cite{a74} modeling the image distribution of past tasks~\cite{a24,a25,a26}.
Nevertheless, they suffer from class imbalance problem due to limited buffer size, or are disabled when data privacy matters.
Dynamic architecture methods~\cite{a16,a17,a18,a19} dynamically create modules  tailored to each task to encode task-specific knowledge.
Although they achieve promising performance in mitigating  catastrophic forgetting, the drastic increasing number of parameters hinders its application in real-world scenarios.
Moreover, these methods fine-tune all the parameters of (sub-)model which are prone to suffer from catastrophic forgetting or increasing computational cost.
Different from the  above-mentioned  methods, the proposed H-Prompts only updates  a small number of learnable prompt parameters over the frozen backbone, effectively reducing both computational costs and the level of catastrophic forgetting associated with past task knowledge.

\begin{figure*}
  \begin{center}
  {\includegraphics[width=17cm, height=9.0cm]{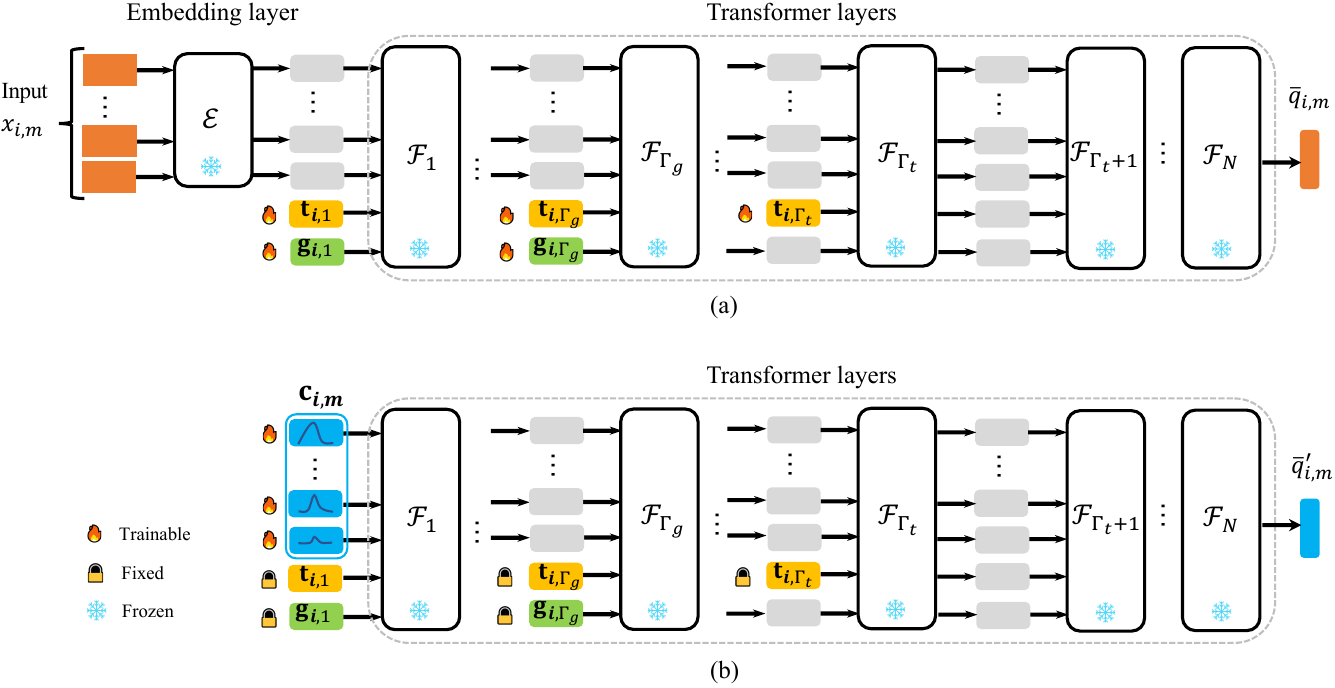}}
  \end{center}
  \caption{An overview of class prompt $\textcolor{blue}{\mathbf{c}_{i,m}}$, task prompt $\textcolor{orange}{\mathbf{t}_i = \{\mathbf{t}_{i,l}\}_{l=1}^{\varGamma_t}}$, and general prompt $\textcolor{green}{\mathbf{g}_i = \{\mathbf{g}_{i,l}\}_{l=1}^{\varGamma_g}}$. (a) depicts trainable task prompt $\mathbf{t}_i = \{\mathbf{t}_{i,l}\}_{l=1}^{\varGamma_t}$ and  general prompt $\mathbf{g}_i = \{\mathbf{g}_{i,l}\}_{l=1}^{\varGamma_g}$  are extended  to the inputs  of frozed multiple Transformer for capturing task knowledge and learning generalized knowledge, respectively.  (b) presents trainable class prompt $\mathbf{c}_{i,m}$ replaces the position of input with fixed task prompt $\mathbf{t}_i = \{\mathbf{t}_{i,l}\}_{l=1}^{\varGamma_t}$ and  general prompt $\mathbf{g}_i = \{\mathbf{g}_{i,l}\}_{l=1}^{\varGamma_g}$ to preserve the knowledge in each class.}
  \label{illustration}
  \end{figure*}

\textbf{Prompt-based Methods.}
Recently, prompt-based methods~\cite{a37,a38,a40} have been paid more attention due to its marvelous performance and  negligible learnable parameters.
Drawing inspiration from  prompt tuning~\cite{a47,a48,a49,a50,Yao_2023_CVPR}, prompt-based methods~\cite{a20,a21,a43} utilize slight learnable prompts to encode task knowledge while keeping backbone frozen to reduce catastrophic forgetting.
For instance,  L2P~\cite{a20} maintains a  pool of prompts and implements instance-level prompt learning in each task.
It selects a subset of prompts from prompt pool for each image to facilitate supervised learning, thereby increasing the likelihood of similar images sharing prompts.
Dualprompt~\cite{a21} employs G-Prompt and E-Prompt to encode task-specific knowledge and task-sharing knowledge, respectively.
E-Prompt is specialized for the knowledge of each task independently and G-Prompt is shared among all the tasks.
Dytox~\cite{a43}  use  prompt to design task-attention block in  the decoder of transformer for modeling the knolwedge of different tasks.
Besides, BIRT~\cite{jeeveswaran2023birt} and ESN~\cite{wang2022isolation} regard the learnable classifer as a prompt and focus on the optimization concept of the classifier.
BIRT~\cite{jeeveswaran2023birt} integrates productive disturbances at different phases of the vision transformer, ensuring consistent predictions in relation to an exponential moving average of the classifier.
ESN~\cite{wang2022isolation} presents a temperature-regulated energy metric, used to denote the confidence score tiers of the task classifier. Moreover, it advocates for a strategy of anchor-based energy self-normalization to guarantee that the task classifier operates within the same energy bracket.
However,  the  prompts of  the above methods only concentrate on the knowledge derived from current task data,  while  ignoring  past task knowledge, resulting in  inferior performance on past tasks.
For solving the above problem,  the proposed H-Prompts  utilizes class prompt to model the distribution of  classes in each task and replayes the past knowledge to  the learning of incoming task with task prompt.
Furthermore,  H-Prompts  adopts self-supervised learning to learn generalized representations in general prompt.

\subsection{Bayesian Neural Networks}

Plain feedforward neural networks are trained with maximum likelihood procedures to obtain the optimal point estimate for all the parameters.However, these networks are prone to overconfidence in their predictions and are incapable of estimating uncertainty.
Bayesian Neural Networks (BNNs) offer a solution to these limitations by representing each parameter as a probability distribution~\cite{a24, a25}. This makes them more robust to perturbations, and the resulting perturbed predictions of BNNs can be employed to model uncertainty~\cite{a26,a27}.
In recent years, BNNs have seen wide-ranging applications in both computer vision and natural language processing domains. For instance, STAR~\cite{a1} utilizes a stoachstic classifer to align the source and target domain data distributions for domain adaptation. DistributionNet~\cite{a54} uses a unique approach where each person's image is modelled as a Gaussian distribution, using the variance to represent image uncertainty for person re-identification.
In the field of natural language processing, BNNs have shown promise as well. For instance, BRNN~\cite{fortunato2017bayesian} introduces a method of incorporating BNNs into recurrent neural networks (RNNs), which are commonly used in NLP tasks, providing a mechanism to capture uncertainty over sequences of data. BAN~\cite{miok2022ban} presents a Bayesian approach that employs Monte Carlo dropout within attention layers of transformer models, providing reliable and well-calibrated estimates for hate speech detection.
However, there is an evident lack of methods that explore the use of BNNs as prompts to characterize data distributions of datasets and evaluate their effectiveness.
Inspired by the rich representation capability of BNNs, we introduce Bayesian Distribution Alignment for leveraging BNNs to model class distributions, serving as class prompt,  and replay the past knowledge during the learning of incoming tasks. Our empirical results reveal that the obtained class prompt effectively captures the data distribution of historical tasks, and successfully mitigates the issue of catastrophic forgetting, a prevalent challenge in continual learning.

\section{Hierarchical Prompts}
In the context of continual learning,  a model is expected to learn the knowledge of non-stationary data from a sequence of tasks, represented as $\{\mathcal{T}_i\}_{i=1}^{T}$, where $T$ refers to the total number of tasks. 
The $i$-th task, denoted by $\mathcal{T}_i = \{(x_{n,i}, y_{n,i})\}_{n=1}^{|\mathcal{T}_i|}$, comprises an image $x_{n,i}\in \mathcal{X}_i$ and its corresponding label $y_{n,i} \in \mathcal{Y}_i$. Here, $\mathcal{X}_i$ and $\mathcal{Y}_i$ symbolize the sets of images and labels in the $i$-th task, respectively. 
Note that the ground-truth label sets are disjoint for different tasks,  $i.e.$,  $\mathcal{Y}_i \cap \mathcal{Y}_j = \varnothing$, if $i \ne j$. 
Moreover, the tasks can be accessed sequentially, and previous tasks are not accessible while learning the current task.
In this work, we focus on challenging exemplar-free class incremental setting, in which  the count of exemplars  from past tasks is zero during training,  and  the task identity is inaccessible during inference.

\begin{figure*}
  \begin{center}
  {\includegraphics[width=18.5cm, height=8.1cm]{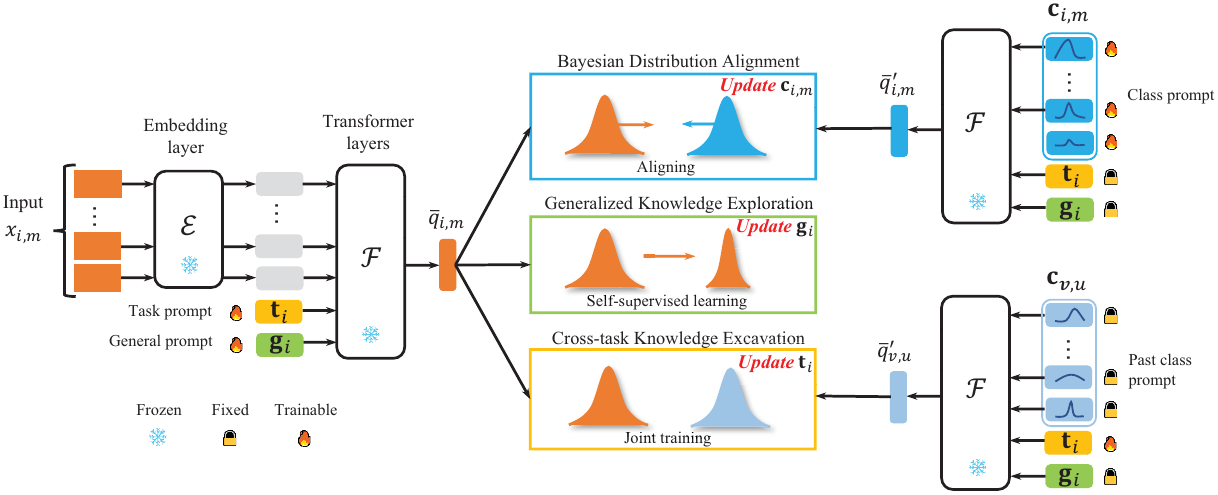}}
  \end{center}
  \caption{
  The framework of the proposed  H-Prompts.
  In the current task, input $x_{i,m}$ (class prompt $\mathbf{c}_{i,m}$)  is extended with task prompt $\mathbf{t}_{i}$ and general prompt $\mathbf{g}_{i}$ to obtain adapted representation $\bar{\mathbf{q}}_{i,m}$  (virtual representation $\bar{\mathbf{q}}'_{i,m}$ ).
  Similarly, we gain past virtual representation $\bar{\mathbf{q}}'_{v,u}$ for past class prompt $\mathbf{c}_{v,u}$. 
  Bayesian Distribution Alignment adjusts $\mathbf{c}_{i,m}$ to  align the distributions between $\bar{\mathbf{q}}_{i,m}$ and $\bar{\mathbf{q}}'_{i,m}$.
  Cross-task Knowledge Excavation  jointly trains   $\bar{\mathbf{q}}'_{v,u}$  and $\bar{\mathbf{q}}_{i,m}$ to optimize $\mathbf{t}_{i}$.
  Moreover, Generalized Knowledge Exploration utilizes $\bar{\mathbf{q}}_{i,m}$ to conduct self-supervised learning for updating  $\mathbf{g_i}$.
  }
  \label{fig2}
  \end{figure*}

\subsection{Overview}

Hierarchical Prompts (H-Prompts) consists of class prompt, task prompt, and general prompt to preserve the knowledge in each class, capture the past task knowledge as well as current task knowledge, and learn generalized  knowledge, respectively.
The overview of class prompt, task prompt, and general prompt is  shown in Figure~\ref{illustration}.
Similar to\cite{a20,a21}, we adopt the frozen vision transformer (ViT)~\cite{a23} as  backbone.
Given an image  $x_{i,m}$  in the $m$-th class of  $i$-th task $\mathcal{T}_{i,m}$,   ViT first divides it into $L$ fixed-sized patches.
The pre-trained embedding layer $\mathcal{E}$ then maps them into a sequence embedding  $\mathbf{f}_{i,m}  \in  \mathbb{R} ^{L \times D}$, where $L$ signifies the sequence length and $D$ refers to the embedding dimension. 
Subsequently, the embedding $\mathbf{f}_{i,m}$ is passed through a stack of Transformer layers $\mathcal{F} = \{\mathcal{F}_l\}_{l=1}^{N}$ to acquire the representation $\mathbf{q}_{i,m} = \mathcal{F}(\mathbf{f}_{i,m}) \in \mathbb{R}^{D}$.\footnote{For simplicity, class token and position token are omitted in this description.}
To obtain the adapted representation specialized for the task, H-Prompts  combines task prompt  $\textcolor{orange}{\mathbf{t}_{i,1}} \in \mathbb{R}^{L_t \times D}$ with general prompt $\textcolor{green}{\mathbf{g}_{i,1}} \in \mathbb{R}^{L_g \times D}$  to attach with the embedding $\mathbf{f}_{i,m}$.
This results in the extended input  $\bar{\mathbf{f}}_{i,m,1} \in \mathbb{R}^{(L_g + L_t + L) \times D}$ of the first Transformer layer $\mathcal{F}_1$:
\begin{align}
  \bar{\mathbf{f}}_{i,m,1} &= [\textcolor{green}{\mathbf{g}_{i,1}}, \textcolor{orange}{\mathbf{t}_{i,1}}, \mathbf{f}_{i,m}], \notag\\
  [\hat{\mathbf{g}}_{i,m,2}, \hat{\mathbf{t}}_{i,m,2} , \hat{\mathbf{f}}_{i,m,2}] &= \mathcal{F}_1(\bar{\mathbf{f}}_{i,m,1}), 
   \label{Eq1}
\end{align}
where $\hat{\mathbf{g}}_{i,m,2}$, $\hat{\mathbf{t}}_{i,m,2}$,  and $\hat{\mathbf{f}}_{i,m,2}$ represent  the outputs of $\mathcal{F}_1$ corresponding with the position of  general prompt, task prompt, and sequence embedding, respectively. The symbols $L_g$ and $L_t$ depict the length of general prompt and task prompt, respectively. 
Moreover, we depict $\hat{\mathbf{g}}_{i,m,l+1}$, $\hat{\mathbf{t}}_{i,m,l+1}$,  and $\hat{\mathbf{f}}_{i,m,l+1}$ as  the outputs of $\mathcal{F}_l$ corresponding with the position of  general prompt, task prompt, and sequence embedding, respectively.
Since different Transformer layers  encode the knowledge in various level,  we extend  task prompt and general prompt to the inputs  of multiple Transformer layers for obtaining superior task knowledge and general knowledge, respectively.
We denote the number of Transformer layers extended with task/general prompt as  $\varGamma_t$/$\varGamma_g$.
For the following $\varGamma_t$/$\varGamma_g$ Transformer layers, 
we replace  $\hat{\mathbf{t}}_{i,m,l}$/$\hat{\mathbf{g}}_{i,m,l}$ in the input of $\mathcal{F}_{l}$  with task/general prompt $\textcolor{orange}{\mathbf{t}_{i,l}}$/$\textcolor{green}{\mathbf{g}_{i,l}}$ to obtain extended input $\bar{\mathbf{f}}_{i,m,l}$:
\begin{align}
  \bar{\mathbf{f}}_{i,m,l} &= [\textcolor{green}{\mathbf{g}_{i,l}}, \textcolor{orange}{\mathbf{t}_{i,l}}, \hat{\mathbf{f}}_{i,m,l}], \notag \\
  [\hat{\mathbf{g}}_{i,m,l+1}, \hat{\mathbf{t}}_{i,m,l+1} , \hat{\mathbf{f}}_{i,m,l+1}] &= \mathcal{F}_{l}(\bar{\mathbf{f}}_{i,m,l}), \notag \\ l &= 1,2,\cdots, \min\{\varGamma_t, \varGamma_g\}
  \label{Eq2}
\end{align}
Assuming that $\varGamma_t > \varGamma_g$ for simplicity, the extended input $\bar{\mathbf{f}}_{i,m,l}$ can be presented as:
\begin{align}
  \bar{\mathbf{f}}_{i,m,l} &= [\hat{\mathbf{g}}_{i,m,l}, \textcolor{orange}{\mathbf{t}_{i,l}}, \hat{\mathbf{f}}_{i,m,l}], \notag \\
  [\hat{\mathbf{g}}_{i,m,l+1}, \hat{\mathbf{t}}_{i,m,l+1} , \hat{\mathbf{f}}_{i,m,l+1}] &= \mathcal{F}_{l}(\bar{\mathbf{f}}_{i,m,l}), \notag \\ l &= \varGamma_g+1,\varGamma_g+2, \cdots, \varGamma_t
  \label{Eq3}
\end{align}
For the last Transformer layers, we have
  \begin{align}
    [\hat{\mathbf{g}}_{i,m,l+1}, \hat{\mathbf{t}}_{i,m,l+1} , \hat{\mathbf{f}}_{i,m,l+1}] &= \mathcal{F}_{l}([\hat{\mathbf{g}}_{i,m,l}, \hat{\mathbf{t}}_{i,m,l} , \hat{\mathbf{f}}_{i,m,l}]) \notag \\ \quad l &= \varGamma_t+1,\varGamma_t + 2, \cdots, N  
  \label{Eq4}
\end{align}
We denote the first token of the outputs of $\mathcal{F}_N$ corresponding with the position of  general prompt  as the adapted representation $\bar{q}$.

For the current $i$-th task, the key issue of H-Prompts  is how to infer the task prompt $\mathbf{t}_i = \{\mathbf{t}_{i,l}\}_{l=1}^{\varGamma_t}$ and the general prompt $\mathbf{g}_i = \{\mathbf{g}_{i,l}\}_{l=1}^{\varGamma_g}$.
The task prompt $\mathbf{t}_i$ should not only encapsulate the discriminative knowledge of current task, but also remember the historical knowledge of past tasks. This requirement can be formally expressed as follows:
\begin{equation}
\mathbf{t}_i \Leftarrow \{\mathcal{T}_i, \mathbb{K}_{i-1}\},
\label{eq:tp}
\end{equation}
where $\mathcal{T}_i$ denotes  the dataset of current task, and $\mathbb{K}_{i-1}$ represents the historical knowledge of all previous $i-1$ tasks:
\begin{equation}
\mathbb{K}_{i-1}=\{\mathbf{k}_{1,1},\mathbf{k}_{1,2},.....,\mathbf{k}_{i-1, |\mathcal{Y}_{i-1}|}\},
\label{eq:tp3}
\end{equation}
where $\mathbf{k}_{i,m}$ signifies the knowledge of the $m$-th class of $i$-th task $\mathcal{T}_{i,m}$.

To derive the task prompt from Eq.~\eqref{eq:tp}, two critical problems arise: 1) the acquisition of the historical knowledge $\mathbb{K}_{i-1}$, and 2) the transference of the historical knowledge $\mathbb{K}_{i-1}$ into the current task prompt $\mathbf{t}_i$.
To store  historical knowledge in a rehearsal-free way,
we introduce Bayesian Distribution Alignment to model the distribution of classes in each task with Bayesian Neural Networks (BNNs), \emph{i.e.,} a multivariate Gaussian distribution $\mathcal{N}(\mathbf{\mu}_{i,m},\mathbf{\Sigma}_{i,m})$ for the $m$-th class of $i$-th task with   learnable mean vector $\mathbf{\mu}_{i,m} \in  \mathbb{R} ^{L \times D}$  and  diagonal covariance matrix $\mathbf{\Sigma}_{i,m} \in  \mathbb{R} ^{L \times D}$.
For prompt tuning, the learnable parameters $\mathbf{\mu}_{i,m}$ and $\mathbf{\Sigma}_{i,m}$ are defined as \textbf{class prompt}  $\textcolor{blue}{\mathbf{c}_{i,m}}$ for optimization, such that
\begin{equation}
  \textcolor{blue}{\mathbf{c}_{i,m}} =  \mathcal{N}(\mathbf{\mu}_{i,m},\mathbf{\Sigma}_{i,m}).
\end{equation}
The details about how to learn  class prompt $\mathbf{c}$  are discussed in Sec.~\ref{Class Prompt}.

Furthermore, we propose Cross-task Knowledge Excavation to transfer the historical class prompts (knowledge) $\{\mathbf{c}_{v,u}|v\in\{1,2,\cdots,i-1\}, u\in\{1,2,\cdots, |\mathcal{Y}_v|\}\}$ and the  knowledge of current dataset $\mathcal{T}_i$ into \textbf{task prompt} $\mathbf{t}_i$.
Task prompt thereby memorizes joint knowledge from past and current tasks (Sec.\ref{Task Prompt}).
To obtain highly generalized  knowledge, we present Generalized Knowledge Exploration to apply  self-supervised learning for obtaining  \textbf{general prompt} $\mathbf{g}_i$  (Sec.~\ref{General Prompt}).

Therefore, class prompt $\mathbf{c}$, task prompt $\mathbf{t}$, and general prompt $\mathbf{g}$ form a novel  prompt-based  paradigm for continual learning  named \textbf{Hierarchical Prompts (H-Prompts)}. 
The framework of H-Prompts is shown in Figure~\ref{fig2}.
The specifics of Bayesian Distribution Alignment, Cross-task Knowledge Excavation, and Generalized Knowledge Exploration are elaborated upon in the subsequent sections, in which we assume for simplicity that task prompts $\mathbf{t}_i$ and general prompts $\mathbf{g}_i$ are extended to the input of the first Transformer layer.

\subsection{Bayesian Distribution Alignment}
\label{Class Prompt}
\begin{figure}
  \begin{center}
  {\includegraphics[width=8.8cm, height=5.2cm]{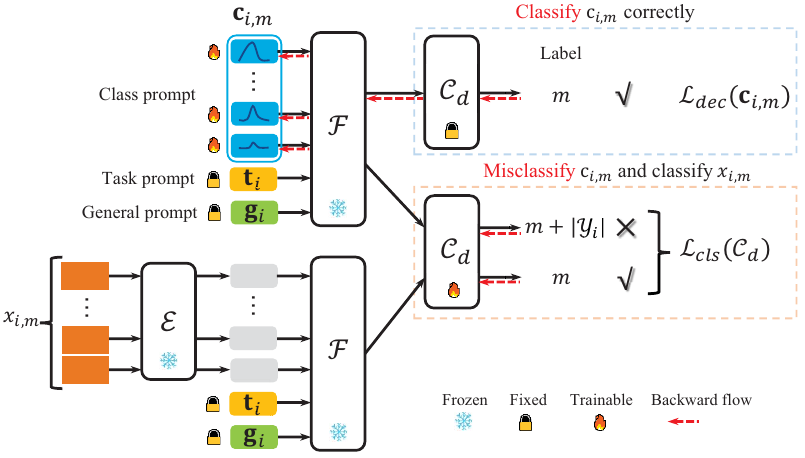}}
  \end{center}
  \caption{
    In Bayesian Distribution Alignment, we first fix discriminative classifier $\mathcal{C}_d$ and update class prompt $\mathbf{c}_{i,m}$ to deceive $\mathcal{C}_d$ by  classifying  $\mathbf{c}_{i,m}$ correctly  with true label $m$. 
    Then, we  fix the  $\mathbf{c}_{i,m}$ and  update the  $\mathcal{C}_d$ to misclassify $\mathbf{c}_{i,m}$ with fake label $m+|\mathcal{Y}_i|$ and  classify input $x_{i,m}$ with true label $m$.
  }
  \label{bda}
  \end{figure}

In this work, we utilize Bayesian Distribution Alignment (BDA) to characterize the data distribution for each class, \emph{e.g.,} class prompt $\mathbf{c}_{i,m} = \mathcal{N}(\mathbf{\mu}_{i,m},\mathbf{\Sigma}_{i,m})$ models the data distribution in the $m$-th class of the $i$-th task $\mathcal{T}_{i,m}$ with learnable parameters $\mathbf{\mu}_{i,m}$ and $\mathbf{\Sigma}_{i,m}$, shown in Figure~\ref{bda}.
By sampling an image $x_{i,m}$ from current dataset $\mathcal{T}_{i,m}$, we define its embedding as $\mathbf{f}_{i,m}$.
By combining fixed task prompt $\mathbf{t}_i$ and  general prompt $\mathbf{g}_i$, we can obtain the extended embedding of  the image $x_{i,m}$ as $\bar{\mathbf{f}}_{i,m} = [\mathbf{g}_i, \mathbf{t}_i, \mathbf{f}_{i,m}]$.
Concurrently, we represent $\mathbf{f}'_{i,m} \in  \mathbb{R} ^{L \times D} $ as  the  virtual embedding   sampled from  class prompt $\mathbf{c}_{i,m}$:
\begin{equation}
\mathbf{f}'_{i,m} \sim  \mathbf{c}_{i,m}=\mathcal{N}(\mathbf{\mu}_{i,m},\mathbf{\Sigma}_{i,m}).
\label{current-classprompt}
\end{equation}

Utilizing $\mathbf{f}'_{i,m}$, we formulate the corresponding extended virtual embedding as $\bar{\mathbf{f}}'_{i,m} = [\mathbf{g}_i, \mathbf{t}_i, \mathbf{f}'_{i,m}]$.
BDA is trained by aligning the distributions of the extended virtual embedding $\bar{\mathbf{f}}'_{i,m}$ and the extended embedding $\bar{\mathbf{f}}_{i,m}$ in an adversarial manner.

In particular, we  leverage a discriminative classifier $\mathcal{C}_d :\mathbb{R}^{D}\to \mathbb{R}^{2|\mathcal{Y}_i|}$ to differentiate between the representations $\bar{\mathbf{f}}_{i,m}$ and $\bar{\mathbf{f}}'_{i,m}$.
The objective of the discriminative classifier $\mathcal{C}_d$ is to accurately classify the real representation $\bar{\mathbf{f}}_{i,m}$ with the correct label $m$, while attempting to misclassify the virtual representation $\bar{\mathbf{f}}'_{i,m}$ with a false label $m+|\mathcal{Y}_i|$:
\begin{gather}
 \mathcal{L}_{cls} (\mathcal{C}_d) = \mathbb{E}_{m,x_{i,m}\in \mathcal{T}_{i,m}}[\ell(m,\mathcal{C}_d(\mathcal{F}(\bar{\mathbf{f}}_{i,m})))] \notag \\+  \mathbb{E}_{m,\mathbf{f}'_{i,m} \sim \mathbf{c}_{i,m}}[\ell(m+|\mathcal{Y}_i|,   \mathcal{C}_d(\mathcal{F}(\bar{\mathbf{f}}'_{i,m})))],
 \label{loss-cls}
\end{gather}
where  $\ell(\cdot, \cdot)$ denotes the cross-entropy loss.
Additionally, we enforce the weights of different classes in $\mathcal{C}_d$ to be orthogonal to each other, thus enhancing the discriminative ability of the classifier. 
Specifically, we define the class weight matrix as $\mathbf{W}  = [\mathbf{w}_1,\mathbf{w}_2,\dots,\mathbf{w}_{2|\mathcal{Y}_i|}] \in \mathbb{R}^{2|\mathcal{Y}_i| \times D}$, and the identity matrix as $\mathbf{E} \in \mathbb{R}^{2|\mathcal{Y}_i| \times 2|\mathcal{Y}_i|} $, where $\mathbf{w}_i$ is the weight of the $i$-th class in $\mathcal{C}_d$.
The discrimination loss is formulated as:
\begin{align}
\mathcal{L}_{dis} (\mathcal{C}_d) &= \ell_M(\mathbf{W}^T\mathbf{W}, \mathbf{E}),
\label{loss-dis}
\end{align}
where $\ell_M(\cdot, \cdot)$ denotes the mean-squared loss.

Furthermore,  the virtual embedding  $\mathbf{f}'_{i,m}$ sampled from  class prompt $\mathbf{c}_{i,m}$ should be optimized to deceive the discriminative classifier $\mathcal{C}_d$, \emph{i.e.},  make $\mathcal{C}_d$  classify virtual representation of  $\bar{\mathbf{f}}'_{i,m}$ correctly:
\begin{gather}
\mathcal{L}_{dec} (\mathbf{c}_{i,m}) =  \mathbb{E}_{m,\mathbf{f}'_{i,m} \sim \mathbf{c}_{i,m}}[\ell(m,\mathcal{C}_d(\mathcal{F}(\bar{\mathbf{f}}'_{i,m})))].
\label{loss-dec}
\end{gather}

The  loss function of BDA is
\begin{align}
 L_{bda}(\mathbf{c}_{i,m}, \mathcal{C}_d) = \mathcal{L}_{cls}(\mathcal{C}_d) + \mathcal{L}_{dis}(\mathcal{C}_d) + \mathcal{L}_{dec}(\mathbf{c}_{i,m}).
\end{align}

Finally, though adversarial training between class prompt $\mathbf{c}_{i,m}$ and discriminative classifier $\mathcal{C}_d$, the class distribution of $m$-th class of $i$-th task can be modeled by $\mathbf{c}_{i,m}=\mathcal{N}(\mathbf{\mu}_{i,m},\mathbf{\Sigma}_{i,m})$, which is stored and and used for future retrieval of the corresponding class knowledge.

\subsection{Cross-task Knowledge Excavation}
\label{Task Prompt}
\begin{figure}
  \begin{center}
  {\includegraphics[width=8.8cm, height=5.5cm]{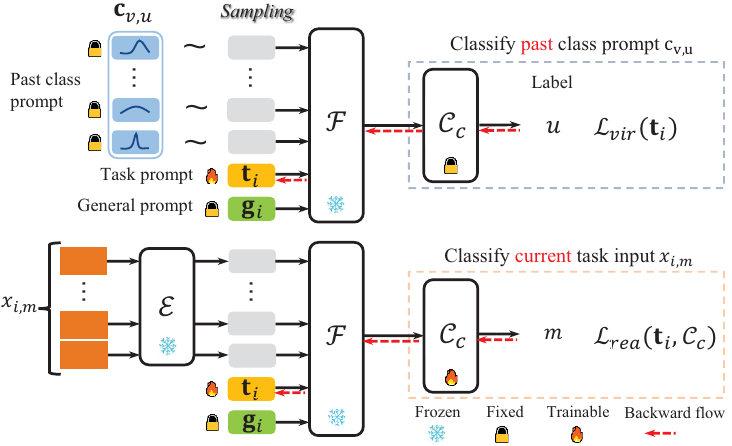}}
  \end{center}
  \caption{In Cross-task Knowledge Excavation, we optimize task prompt $\mathbf{t}_{i}$ to classify the input sampled from past class prompt $\mathbf{c}_{v,u}$ for encoding past task knowledge. 
  Moreover, we  update  $\mathbf{t}_{i}$ and classification classifier $\mathcal{C}_{c}$ to classify input $x_{i,m}$ for learning  current task knowledge.}
  \label{cke}
  \end{figure}

Leveraging the obtained class prompts of  historical classes, we propose Cross-task Knowledge Excavation (CKE) to integrate the  knowledge of current task with  past  knowledge  replayed by historical  class prompts for task prompt,  as shown in  Figure~\ref{cke}.
For the current $i$-th task, CKE guides the task prompt $\mathbf{t}_i$ to assimilate knowledge from the embedding $\{\mathbf{f}_{i,m}\}_{m=1}^{|\mathcal{Y}_i|}$ of all $|\mathcal{Y}i|$ classes, along with the virtual embedding $\{\mathbf{f}'_{v,u}| v\in\{1,2,\cdots, i-1\}, u\in\{1,2,\cdots, |\mathcal{Y}_v|\}\}$ of all classes from the previous $i-1$ tasks.
The virtual embedding  $\mathbf{f}'_{v,u}$ of the  $u$-th class in $v$-th task is  replayed by its corresponding class prompt $\mathbf{c}_{v,u}$:
\begin{align}
\mathbf{f}'_{v,u} \sim \mathbf{c}_{v,u}=\mathcal{N}(\mathbf{\mu}_{v,u},\mathbf{\Sigma}_{v,u}).
\label{past-classprompt}
\end{align}

With the corresponding general prompt $\mathbf{g}_i$ and  task prompt $\mathbf{t}_{i}$, we can obtain the extended embedding $\bar{\mathbf{f}}_{i,m} = [\mathbf{g}_i, \mathbf{t}_{i}, \mathbf{f}_{i,m}]$ and extended virtual embedding  $\bar{\mathbf{f}}'_{v,u} = [\mathbf{g}_{i}, \mathbf{t}_{i}, \mathbf{f}'_{v,u}]$ of past classes.
For the current task, we optimize the task prompt $\mathbf{t}_{i}$ and the classification classifier $\mathcal{C}_{c}$ to  classify the extended embedding $\bar{\mathbf{f}}_{i,m}$:
\begin{gather}
 \mathcal{L}_{rea}(\mathbf{t}_i, \mathcal{C}_c) = \mathbb{E}_{m,x_{i,m}\in \mathcal{T}_{i,m}}[\ell(m,\mathcal{C}_c(\mathcal{F}(\bar{\mathbf{f}}_{i,m})))].
\end{gather}

In addition, for encoding the past task knowledge, we optimize the task prompt $\mathbf{t}_{i}$ to accurately classify extended virtual embedding  from  past $i-1$ tasks:
\begin{gather}
 \mathcal{L}_{vir}(\mathbf{t}_i) =  \mathbbm{1}_{[i > 1]}  \mathbb{E}_{v,u,\mathbf{f}'_{v,u} \sim \mathbf{c}_{v,u}}[\ell(u,\mathcal{C}_c(\mathcal{F}(\bar{\mathbf{f}}'_{v,u})))],
\end{gather}
where $\mathbbm{1}_{[i > 1]}$ is an indicator function, with $\mathbbm{1}_{[i > 1]} = 1$ if $i > 1$, and $\mathbbm{1}_{[i > 1]} = 0$ otherwise.

The loss function of CKE is:
\begin{gather}
 \mathcal{L}_{cke}(\mathbf{t}_i, \mathcal{C}_c) = \mathcal{L}_{rea}(\mathbf{t}_i, \mathcal{C}_c) + \lambda \mathcal{L}_{vir}(\mathbf{t}_i),
\label{loss-cke}
\end{gather}
where $\lambda$ acts as a trade-off parameter, typically defaulted to 0.1.

\subsection{Generalized Knowledge Exploration}
\label{General Prompt}

The representation of current task learned  by class prompt and task prompt with supervised learning may be sub-optimal for other tasks. 
Therefore,  we present Generalized Knowledge Exploration (GKE) for   general prompt to learn  generalized  representation via self-supervised learning with pretext tasks.
Drawing inspiration from SupContrast~\cite{a29}, which employs class information for self-supervised training to learn class discriminative representation promoting generalized discriminative representation learning, GKE adopts class discrimination as the pretext task.  
Consider a set of $N$ random sampled images  in the $i$-th task, we perform two sets of augmentation for each image and obtain $2N$ images $\{x_{i,n}\}_{n=1}^{2N}$, the embeddings of which  are $\{\mathbf{f}_{i,n}\}_{n=1}^{2N}$.
The extended embedding and adapted representation are denoted as $\bar{\mathbf{f}}_{i,n} = [\mathbf{g}_{i}, \mathbf{t}_{i}, \mathbf{f}_{i,n}]$ and $\bar{\mathbf{q}}_{i,n} = \mathcal{F}(\bar{\mathbf{f}}_{i,n})$, respectively.
We leverage label information to ensure that representations from the same class are clustered together while representations from different classes are dispersed:
\begin{align}
& \mathcal{L}_{gke}(\mathbf{g}_i) = \notag \\ &\mathbb{E}_{x_{n,i}\in \mathcal{T}_i}[\frac{-1}{\mathcal{P}(x_{n,i})} \sum_{p \in \mathcal{P}(x_{n,i})} \log \frac{e^{d(\bar{\mathbf{q}}_{n,i},\bar{\mathbf{q}}_{p,i})/\tau}}{\sum_{k=1}^{2N}\mathbbm{1}_{[k \ne n]} e^{d(\bar{\mathbf{q}}_{n,i},\bar{\mathbf{q}}_{k,i})/\tau}}],
\label{loss-gke}
\end{align}
where $\mathcal{P}(x_{n,i})$ represents the postive images sharing  the same label as $x_{n,i}$, $d$ depicts the cosine similarity, and $\tau$   is a scalar temperature parameter.

\begin{algorithm*}[t]
  \caption{Training process of  H-Prompts} 
  \label{train}
  \hspace*{0.02in} {\bf Input:} 
  Training set $\{\mathcal{T}_i\}_{i=1}^{T}$, pre-trained embedding layer $\mathcal{E}$, pre-trained Transformer layers $\mathcal{F}$, task prompts $\{\mathbf{t}_{i} = \{\mathbf{t}_{i,l}\}_{l=1}^{\varGamma_t}\}_{i=1}^{T} $, general prompts $\{\mathbf{g}_{i} = \{\mathbf{g}_{i,l}\}_{l=1}^{\varGamma_g}\}_{i=1}^{T} $, class prompts $\{\{\mathbf{c}_{i,m}\}_{m=1}^{|\mathcal{Y}_i|}\}_{i=1}^{T}$, discriminative classifier $\mathcal{C}_d$, classification classifier $\mathcal{C}_c$,  GKE epoch $E_{gke}$, total epoch $E_{max}$; \\
  \hspace*{0.02in} {\bf Output:} 
  $\{\mathbf{t}_i\}_{i=1}^{T}$, $\{\mathbf{g}_i\}_{i=1}^{T}$, $\mathcal{C}_c$, $\{K_i\}_{i=1}^T$;
  \begin{algorithmic}[1]
 
  \For{ all task $i$ = 1, 2, $\dots$, $T$}
  \State  Initialize $\mathbf{g}_i$,  $\mathbf{t}_i$, $\{\mathbf{c}_{i,m}\}_{m=1}^{|\mathcal{Y}_i|}$, $\mathcal{C}_d$, $\mathcal{C}_c$;
  \For{$\Omega$ = 1, 2, 3, $\dots$, $E_{gke}$}
  \State // \textbf{Generalized Knowledge Exploration}

  \State Fetch a mini-batch $B$ from $\mathcal{T}_i$;
  \State Obtain  sequence embedding $\mathbf{f}_i$ of  each image in $B$ with $\mathcal{E}$;
  \State Obtain the extended inputs  $\bar{\mathbf{f}}_i$ with $\mathbf{g}_i$ and   $\mathbf{t}_i$;
  \State Optimize $\mathbf{g}_i$ by   $\min \mathcal{L}_{gke}(\mathbf{g}_i)$ with $\bar{\mathbf{f}}_i$ via Eq.~\ref{loss-gke};
  \EndFor
  \For{$\Omega$ = $E_{gke}$+1, $E_{gke}$+ 2, $\dots$, $E_{max}$}
  \State Fetch a mini-batch $B$ from $\mathcal{T}_i$;
  \State // \textbf{Bayesian Distribution Alignment}
  \State Obtain  sequence embedding $\mathbf{f}_i$ of  each image in $B$ with $\mathcal{E}$;
  \State Obtain the extended inputs $\bar{\mathbf{f}}_i$ with $\mathbf{g}_i$ and   $\mathbf{t}_i$;
  \State Obtain virtual sequence embedding $\mathbf{f}'_i$ sampled  from  $\{\mathbf{c}_{i,m}\}_{m=1}^{|\mathcal{Y}_i|}$;
  \State Obtain the extended virtual inputs $\bar{\mathbf{f}}'_{i}$ with $\mathbf{g}_i$ and   $\mathbf{t}_i$;
  \State Optimize $\mathcal{C}_d$ by  $\min (\mathcal{L}_{cls}(\mathcal{C}_d) + \mathcal{L}_{dis}(\mathcal{C}_d))$ with $\bar{\mathbf{f}}_i$ and $\bar{\mathbf{f}}'_{i}$ via Eq.~\ref{loss-cls} and Eq.~\ref{loss-dis};
  \State Optimize $\{\mathbf{c}_{i,m}\}_{m=1}^{|\mathcal{Y}_i|}$ by  $\min \mathcal{L}_{dec} (\mathbf{c}_{i,m})$ with $\bar{\mathbf{f}}'_{i}$ via Eq.~\ref{loss-dec};
  \State // \textbf{Cross-task Knowledge Excavation}
  \State Obtain virtual sequence embedding $\mathbf{f}'_v$ sampled  from  $\{\{\mathbf{c}_{v,u}\}_{u=1}^{|\mathcal{Y}_v|}\}_{v=1}^{i}$;
  \State Obtain the extended virtual inputs $\bar{\mathbf{f}}'_{v}$  with $\mathbf{g}_i$ and   $\mathbf{t}_i$;
  \State Optimize $\mathbf{t}_i, \mathcal{C}_c$ by  $\min \mathcal{L}_{cke}(\mathbf{t}_i, \mathcal{C}_c)$ with $\bar{\mathbf{f}}_i$ and $\bar{\mathbf{f}}'_{v}$ via Eq.~\ref{loss-cke};
  \EndFor
  \State Adopt $\mathcal{E}$, $\mathcal{F}$ with  $\mathbf{t}_i$,  $\mathbf{\mathbf{g}}_i$ as the query function $q_i(\cdot)$;
  \State Extract the query outputs $Q_i$ of all images in  $\mathcal{T}_i$ with $q_i(\cdot)$;
  \State Obtain  task-aware keys  $K_i = \{\kappa_i^j\}_{j=1}^o$ by  K-Means with $Q_i$;

  \EndFor
  \State \Return $\{\mathbf{t}_i\}_{i=1}^{T} $, $\{\mathbf{g}_i\}_{i=1}^{T}$, $\mathcal{C}_c$, $\{K_i\}_{i=1}^T$.
  \end{algorithmic}
  \end{algorithm*}

  \begin{algorithm*}[t]
    \caption{Testing process of  H-Prompts} 
    \label{test}
    \hspace*{0.02in} {\bf Input:} 
    Test image $x_e$, pre-trained embedding layer $\mathcal{E}$, pre-trained Transformer layers $\mathcal{F}$, task prompts $\{\mathbf{t}_{i} = \{\mathbf{t}_{i,l}\}_{l=1}^{\varGamma_t}\}_{i=1}^{T} $, general prompts $\{\mathbf{g}_{i} = \{\mathbf{g}_{i,l}\}_{l=1}^{\varGamma_g}\}_{i=1}^{T} $,  classification classifier $\mathcal{C}_c$,  task-aware keys  $\{K_i\}_{i=1}^T$, query function $\{q_i(\cdot)\}_{i=1}^{T}$; \\
    \hspace*{0.02in} {\bf Output:} 
    $y_e$;
    \begin{algorithmic}[1]
    \For{ all task $i$ = 1, 2, $\dots$, $T$}
    \State Obtain query feature  $q_i(x_e)$;
    \State Calculate the distance $d_i$ between  $q_i(x_{e})$ and  $K_i = \{\kappa_i^j\}_{j=1}^o$: $d_i =\min_{j \in \{1,\dots, o\}} dis(q_i(x_{e}), \kappa_i^j)$;
    \EndFor
   \State Obtain task identity $\text{id}(x_e)$ for  $x_e$: $\text{id}(x_e) = \arg \min_{i \in \{1,\dots, T\}} d_i$;
   \State Obtain prediction $y_e$ with $\mathcal{E}$, $\mathcal{F}$, $\mathcal{C}_c$, $\{\mathbf{g}_{\text{id}(x_e),l}\}_{l=1}^{\varGamma_g}$, $\{\mathbf{t}_{\text{id}(x_e),l}\}_{l=1}^{\varGamma_t}$;
   \State \Return $y_e$.
    \end{algorithmic}
    \end{algorithm*}

\subsection{Total Objective}
\label{Total Objective}
We amalgamate the above-mentioned class prompt, task prompt, and general prompt into a unified framework, referred to as Hierarchical Prompts (H-Prompts). The total loss of H-Prompts is expressed as follows:
\begin{align}
 \mathcal{L}_{hp} =L_{bda}(\mathbf{c}, \mathcal{C}_d) +  \mathcal{L}_{cke}(\mathbf{t}, \mathcal{C}_c) +  \mathcal{L}_{gke}(\mathbf{g}).
\end{align}
The training  process is illustrated in Algorithm~\ref{train}.

\begin{table*}
\centering
\scriptsize
\setlength{\tabcolsep}{13.4pt}
\footnotesize
\renewcommand{\arraystretch}{1.1}
\caption{Comparison with exemplar-based methods, regularization-based methods, and recent prompt-based methods on Split CIFAR-100 and Split ImageNet-R implemented on VIT-B/16~\cite{a23}. The best performance is highlighted in \textbf{bold}. }
\label{tab1}

{
\begin{tabular}{l|| c| c  c ||c| c  c }
\shline
\multirow{2}{*}{Methods} & \multirow{2}{*}{Buffer size}& \multicolumn{2}{c||}{Split CIFAR-100}& \multirow{2}{*}{Buffer size}&\multicolumn{2}{c} {Split ImageNet-R}\\
 &  & Average Acc $\uparrow$  & Forgetting $\downarrow$ &  & Average Acc $\uparrow$ & Forgetting $\downarrow$ \\
 \hline
ER~\cite{a61} & \multirow{5}{*}{1000} & 67.9$\pm$ 0.6 & 33.3$\pm$1.3 & \multirow{5}{*}{1000} & 55.1$\pm$1.3 & 35.4$\pm$0.5 \\
BiC~\cite{a63} &  & 66.1$\pm$1.8 & 35.2$\pm$1.6 &  & 52.1$\pm$1.1 & 36.7$\pm$1.1 \\
GDumb~\cite{a62} &  & 67.1$\pm$0.4 & - &  & 38.3$\pm$0.6 & - \\
DER++~\cite{a64} &  & 61.1$\pm$0.9 & 39.9$\pm$1.0 &  & 55.5$\pm$1.3 & 34.6$\pm$1.5 \\
C$\textnormal{o}^2$L~\cite{a15} &  & 72.6$\pm$1.3 & 28.6$\pm$1.6 & & 53.5$\pm$1.6 & 37.3 $\pm$ 1.8 \\
\hline
Dytox~\cite{a43} & \multirow{2}{*}{2000} & 68.4$\pm$1.1 & 33.9$\pm$0.8 & \multirow{2}{*}{2000} & - & - \\
BIRT~\cite{jeeveswaran2023birt} &  & 66.7$\pm$0.4 & 19.0$\pm$2.0 &  & - & - \\
\hline
ER~\cite{a61} & \multirow{5}{*}{5000} & 82.5$\pm$0.8 & 16.5$\pm$0.3 & \multirow{5}{*}{5000} & 65.2$\pm$0.4 & 23.3 $\pm$ 0.9 \\
BiC~\cite{a63} &  & 81.4$\pm$0.9 & 17.3$\pm$1.0 &  & 64.6$\pm$1.3 & 22.3$\pm$1.7 \\
GDumb~\cite{a62} &  & 81.7$\pm$0.1 & - &  & 65.9$\pm$0.3 & - \\
DER++~\cite{a64} &  & 83.9$\pm$0.3 & 14.6$\pm$ 0.7 &  & 66.7$\pm$0.9 & 20.7$\pm$1.2 \\
C$\textnormal{o}^2$L~\cite{a15} &  & 82.5$\pm$0.9 & 17.5$\pm$1.8 & & 65.9$\pm$0.1 & 23.4$\pm$0.7 \\
\hline
FT-seq & \multirow{6}{*}{0} & 33.6$\pm$0.9 & 86.9$\pm$0.2 & \multirow{6}{*}{0} & 28.9$\pm$1.4 & 63.8$\pm$1.5 \\
EWC~\cite{a8} &  & 47.0$\pm$0.3 & 33.3$\pm$1.2 &  & 35.0$\pm$0.4 & 56.2$\pm$0.9 \\
LwF~\cite{a60} &  & 60.7$\pm$0.6 & 27.8$\pm$2.2 &  & 38.5$\pm$1.2 & 52.4$\pm$0.6 \\
L2P~\cite{a20} &  & 83.9$\pm$0.3 & 7.4$\pm$0.4 &  & 61.6$\pm$0.7 & 9.7$\pm$0.5 \\
ESN~\cite{wang2022isolation} &  & 86.3$\pm$0.5 & 4.8$\pm$0.1 &  & - & - \\
DualPrompt~\cite{a21} &  & 86.5$\pm$0.3 & 5.2$\pm$0.1 & & 68.1$\pm$0.5 & 4.7$\pm$0.2 \\
\hline
\textbf{H-Prompts}& 0 & \textbf{87.8$\pm$0.3}  & \textbf{4.1$\pm$0.2}  & 0 & \textbf{70.6$\pm$0.3} & \textbf{4.0$\pm$0.2} \\
\hline
Upper-bound& - & 90.9$\pm$0.1 & - & - & 79.1$\pm$0.2 & - \\
\shline
\end{tabular}}
\end{table*}

\subsection{Inference Strategy}
\label{Inference Strategy}
During training, task prompt and general prompt are learned for each task. However, during inference, the task identity of each testing image is not accessible. As a solution, we introduce a novel task-aware query-key mechanism for selecting the task identity for each testing image.

Specifically,  we adopt the pre-trained model with task prompt $\mathbf{\mathbf{t}}_i$ and general prompt $\mathbf{\mathbf{g}}_i$ as the query model $q_i(\cdot)$.
With the query model $q_i(\cdot)$, we extract the query outputs $\{q_i(x_{n,i})\}_{n=1}^{|\mathcal{T}_i|}$ of all images $\{x_{n,i}\}_{n=1}^{|\mathcal{T}_i|}$  at the end of training in the $i$-th task. 
Moreover,  we perform K-means with all query outputs $\{q_i(x_{n,i})\}_{n=1}^{|\mathcal{T}_i|}$   to obtain $o$ task-aware cluster centers (keys)  $K_i = \{\kappa_i^1,\dots,\kappa_i^o\}$ embodying fine-grained semantic knowledge of the $i$-th task,   where $o$ represents  the number of task-aware keys.
Given a testing sample $x_e$ without the task identity,  we first calculate the distance between the query output $q_i(x_{e})$ and task-aware keys $K_i$  under all the task identity $i \in \{1,\dots, T\}$:
\begin{gather}
d_i =\min_{j \in \{1,\dots, o\}} dis(q_i(x_{e}), \kappa_i^j), 
\end{gather}
where $dis$ represents the  Euclidean distance.
Subsequently, we choose the identity with the smallest distance as the task identity of testing sample $x_e$:
\begin{gather}
\text{id}(x_e) = \arg \min_{i \in \{1,\dots, T\}} d_i.
\end{gather}
Having obtained the task identity $\text{id}(x_e)$ for testing sample $x_e$, we select general prompt $\mathbf{g}_{\text{id}(x_e)}$ and task prompt $\mathbf{t}_{\text{id}(x_e)}$ of the $\text{id}(x_e)$-th task with the pre-trained backbone to carry out inference.
The testing  process is depicted in Algorithm~\ref{test}.

\section{Experiments}
\subsection{Settings}

\subsubsection{Datasets} 

We evaluate the proposed method on two stardard class incremental learning benchmarks, \emph{i.e.}, Split CIFAR-100~\cite{a58} and  Split ImageNet-R~\cite{a59}. 
CIFAR-100 is a tiny image recognition dataset consisting of 60,000 32x32 colour images in 100 classes, with 500 training images and 100 testing images per class. 
Split CIFAR-100 is constructed by splitting CIFAR-100 into 10 disjoint tasks, each containing 10 classes per task. 
Although CIFAR-100 is a simple task for image recognition, it is challenge in continual learning scenario.
However,   the training set of pre-trained  vision backbone~\cite{a23} usually contains ImageNet~\cite{a55}, which may lead to information leakage for the continual learning of Split CIFAR-100.
ImageNet-R, an alternative to ImageNet-based benchmark, has a different data distribution from ImageNet.
It comprises 30,000 images of 200 ImageNet object classes in diverse textures and styles, \emph{e.g.}, paintings, sculptures, and embroidery. 
Split ImageNet-R divides ImageNet-R into 10 disjoint tasks, each including 20 classes. Following the strategy of Dualprompt~\cite{a21}, we split ImageNet-R into a training set with 24,000 images and a testing set with 6,000 images.

\subsubsection{Baseline} 

We evaluate the proposed H-Prompts against multiple methods including regularization-based~\cite{a8,a60}, exemplar-based~\cite{a61,a62,a63,a64,a15}, architecture-based~\cite{a65,a66,a67,a68}, and prompt-based methods~\cite{a20,a21,wang2022isolation, a43, jeeveswaran2023birt}. 
Additionally, comparisons are made with FT-seq and Upper-bound, the former represents a naive fine-tuning approach for training sequential tasks, while the latter indicates a method that trains all tasks simultaneously.

\subsubsection{Performance Metrics} 

For the evaluation of regularization-based and rehearsal-based methods, we utilize the average accuracy $A_{T}$ and forgetting rate $F_{T}$. 
The average accuracy is defined as the mean test accuracy of all tasks post completion of the sequential task training. The forgetting rate indicates the average performance decline from the highest previous result to the final result for each task. Formally, average accuracy $A_{T}$ and forgetting rate $F_{T}$ are defined as follows:
\begin{gather}
  A_T = \frac{1}{T}\sum_{i=1}^{T}A_{T,i}, \\
  F_T = \frac{1}{T-1}\sum_{i=1}^{T-1}\max_{j\in {1,\cdots, T-1}}(A_{j,i} - A_{T,i}),
  \label{eq:da}
\end{gather}
where $A_{j,i}$ represents the accuracy of task $\mathcal{T}_j$ subsequent to training on task $\mathcal{T}_i$. Higher average accuracy $A_{T}$ and lower forgetting rate $F_{T}$ indicate superior performance.

Additionally, due to the distinct backbone of H-Prompts  and dynamic architectures methods, a comparison of the average accuracy for classification tasks may not be fair.
Therefore, we compute the difference $\Delta$ between the upper-bound accuracy of the employed backbone and the accuracy of each method to evaluate the performance of dynamic architectures methods, as recommended in Dualprompt~\cite{a21}. Formally, $\Delta$ is expressed as:
\begin{gather}
  \Delta = A_U - A_T,
\end{gather}
where $A_T$ denotes the average accuracy, and $A_U$ depicts the result of upper-bound accuracy,   which is the result of supervised learning on all data with an independent and identically distributed assumption.

\begin{table*}
  \centering
  \scriptsize
  \setlength{\tabcolsep}{18.8pt}
  \footnotesize
  \renewcommand{\arraystretch}{1.1}
  \caption{Comparison with architecture-based methods on Split CIFAR-100. ``Diff.'' depicts the difference between the performance in  each method and upper-bound  of its pre-trained model.  The best performance is highlighted in \textbf{bold}.}
  \label{tab2}
  
  {
  \begin{tabular}{l|| c| c | c c  }
  \shline
  \multirow{2}{*}{Methods} & \multirow{2}{*}{BackBone}&\multirow{2}{*}{Buffer size }& \multirow{2}{*}{Average Acc $\uparrow$ }& \multirow{2}{*}{Diff. $\downarrow$}\\
  &&&&\\
   \hline
  SupSup~\cite{a66} & \multirow{5}{*}{ResNet18}& 0& 28.3$\pm$2.5 & 52.1   \\
  DualNet~\cite{a65} &  & 1000& 40.1$\pm$1.6 & 40.3  \\
  RPSNet~\cite{a68} &  & 2000& 68.6  & 11.8  \\
  DynaER~\cite{a67} &  & 2000& 74.6 & 5.8  \\
  Upper-bound &  & -& 80.4 & -   \\
  
  \hline
  DynaER~\cite{a67} &  \multirow{2}{*}{ResNet152}& 2000& 71.0$\pm$0.6 & 17.5   \\
  Upper-bound &  & -& 88.5 & -   \\
  
  \hline
  L2P~\cite{a20}& \multirow{4}{*}{VIT-B/16} & 0& 83.9$\pm$0.3 & 7.0   \\
  DualPromot~\cite{a21} &  & 0& 86.5$\pm$0.3 & 4.4   \\
  
  \textbf{H-Prompts} &  & 0& \textbf{87.8$\pm$0.3} & \textbf{3.1}   \\
  Upper-bound &  & -& 90.9$\pm$0.1 & -  \\

  \shline
  \end{tabular}}
  \end{table*}

\subsubsection{Implementation Details} 
We use the  VIT-B/16~\cite{a23} pre-trained on ImageNet~\cite{a55} as backbone, whose parameters are all frozen during training.
All images are resized to 224 $\times$ 224 and normalized within the range of [0,1].
For training, we adopt  Adam~\cite{a69} with learning rate of 0.02 and 0.006 for class prompt and task prompt, respectively.
The SGD~\cite{a70} optimizer with learning rate of 0.001 is applied for optimizing the general prompt.
The embedding dimension $D$ is set to 768.
For Split CIFAR-100, we set the GKE epoch $E_{gke}$ to 5 and total epoch $E_{max}$ to 15.
The length of class prompt $L_c$, task prompt $L_t$, and general prompt $L_g$ are 196, 5, and 1, respectively.
The depth of task prompt $D_t$ is 5, and the number of task-aware keys $o$ is 8 per class.
Since the backbone pre-trained on ImageNet struggles to learn discriminative representation in Split ImageNet-R, we employ different experiment settings  to ensure a properly fit to the training data.
Specifically,  the number of total epoch $E_{max}$ is  65,  the length of task prompt $L_t$ is 25, the depth of task prompt $D_t$ is 7, the number of task-aware keys $o$ is 4 per class.

\begin{table}
  \centering
  \scriptsize
  \setlength{\tabcolsep}{7.5pt}
  \footnotesize
  \renewcommand{\arraystretch}{1.2}
  \caption{Ablation study of H-Prompts  on Split CIFAR-100 with one task-aware key per task. The best performance is highlighted in \textbf{bold}.}
  
  \label{model analyses}
  
  {
  \begin{tabular}{c|c c c|c | c}
  \shline
  
  \multirow{2}{*}{} &\multirow{2}{*}{TP} &\multirow{2}{*}{GP}&\multirow{2}{*}{CP}&\multirow{2}{*}{Average Acc $\uparrow$ } &\multirow{2}{*}{Forgetting $\downarrow$}\\
  &&&&&\\
  \hline
  FT-seq &\xmark&\xmark&\xmark& 33.6& 86.9\\
  TP &\cmark&\xmark&\xmark& 85.5& 5.9\\
  TGP &\cmark&\cmark&\xmark& 86.2& 5.5\\
  H-Prompts  &\cmark&\cmark&\cmark& \textbf{86.9}& \textbf{4.7}\\
  
  \shline
  \end{tabular}}
  \label{tab3}
  \end{table}

\begin{table}
\centering
\scriptsize
\setlength{\tabcolsep}{7.5pt}
\footnotesize
\renewcommand{\arraystretch}{1.3}
\caption{Analysis of inference strategy  in H-Prompts  on Split CIFAR-100 with one key per task. The best performance is highlighted in \textbf{bold}.}

{
\begin{tabular}{c|c | c}
\shline
\multirow{2}{*}{Ablated components} &\multirow{2}{*}{Average Acc $\uparrow$} &\multirow{2}{*}{Forgetting $\downarrow$}\\
&&\\
\hline
Naive query-key mechanism & 86.2& 5.0\\
\hline
Task-aware query-key mechanism  & \textbf{86.9}& \textbf{4.7}\\

\shline
\end{tabular}}
\label{tab4}
\end{table}

\subsection{Comparison with Existing Methods}

H-Prompts is evaluated in comparison with traditional exemplar-based methods~\cite{a61,a62,a63,a64,a15}, regularization-based methods~\cite{a8,a60}, and recent prompt-based methods~\cite{a20,a21,wang2022isolation, a43, jeeveswaran2023birt}, as highlighted in Table~\ref{tab1}.
Our evaluation reveals that H-Prompts outperforms these methods on both Split CIFAR-100 and Split ImageNet-R datasets.
Specifically, H-Prompts shows superior performance to the state-of-the-art (SOTA) exemplar-based method DER++~\cite{a64} with a buffer size of 5000, outdoing it by 3.9\% on both datasets.
This results demonstrates the superiority of H-Prompts over exemplar-based methods, \emph{i.e.}, it achieves better performance without requiring a large amount of memory to store historical samples.
H-Prompts also surpasses the best-performing regularization-based method, LwF~\cite{a60} by 27.1\% and 32.1\% on Split CIFAR-100 and Split ImageNet-R datasets, respectively.
This significant performance boost demonstrates the effectiveness of H-Prompts compared to regularization-based methods.
Furthermore, H-Prompts outperforms the SOTA prompt-based method Dualprompt~\cite{a21} by 1.3\% and 2.5\% on Split CIFAR-100 and Split ImageNet-R datasets, respectively, and exceeds the contemporary prompt-based method ESN~\cite{wang2022isolation} by 1.5\% on Split CIFAR-100.
These results prove the effectiveness of our prompt design strategy.
In addition to these achievements, our method exhibits the lowest forgetting rate among all tested methods, further attesting to the effectiveness of H-Prompts

Further comparison is made between H-Prompts and dynamic architecture methods~\cite{a66,a66,a67,a68}, with the results displayed in Table~\ref{tab2}.
When compared to the best-performing dynamic architecture method DynaER~\cite{a67}, with a buffer size of 2000 yielding an average accuracy of 71.0\% and a difference of 17.5\%, H-Prompts demonstrates superior performance, achieving an average accuracy of 87.8\% and a difference of 3.1\%, all without the necessity of a memory buffer. 
Furthermore, when compared to the SOTA Dualprompt~\cite{a21}, H-Prompts improves the average accuracy by 1.3\% and concurrently reduces the difference by 1.3\%, providing further evidence of the effectiveness of H-Prompts.

\subsection{Ablation Studies and Analysis}
\textbf{Ablation study of H-Prompts.}
We conduct an ablation study about the task prompt (TP), general prompt (GP), and class prompt (CP) of the proposed H-Prompts  on Split CIFAR-100, and summarize the results in Table~\ref{tab3}.
To isolate the impact of task-aware key numbers on H-Prompts' performance, a single task-aware key is utilized in each task. The results indicate that the TP model improves average accuracy by 51.5\% and reduces the forgetting rate by 81.0\% compared to FT-seq, thereby emphasizing the efficacy of task prompts. The combination of general prompt and task prompt in the TGP model enhances the average accuracy by 0.7\% and reduces the forgetting rate by 0.4\% relative to TP. This suggests that the general knowledge derived from the general prompt can alleviate catastrophic forgetting. By incorporating the class prompt's historical knowledge into TGP, H-Prompts achieves the highest average accuracy (86.9\%) and the lowest forgetting rate (4.7\%), implying that the class prompt's modeling of past tasks effectively mitigates forgetting of past task knowledge.

\begin{table}[t]
  \centering
  \scriptsize
  \setlength{\tabcolsep}{14.5pt}
  \footnotesize
  \renewcommand{\arraystretch}{1.3}
  \caption{Comparison between H-Prompts and H-Prompts (w. sgp). H-Prompts (w. sgp) denotes H-Prompts with shared general prompt. The best performance is highlighted in \textbf{bold}.}
  
  {
  \begin{tabular}{c|c | c}
  \shline
  \multirow{2}{*}{Methods} &\multirow{2}{*}{Average Acc $\uparrow$} &\multirow{2}{*}{Forgetting $\downarrow$}\\
  &&\\
  \hline
  H-Prompts (w. sgp)& 86.5& 4.7\\
  \hline
  H-Prompts  & \textbf{86.9}& \textbf{4.7}\\
  
  \shline
  \end{tabular}}
  \label{H-Prompts (w. sgp)}
  \end{table}

  \begin{figure}[t]
    \begin{center}
    {\includegraphics[width=8cm, height=4.5cm]{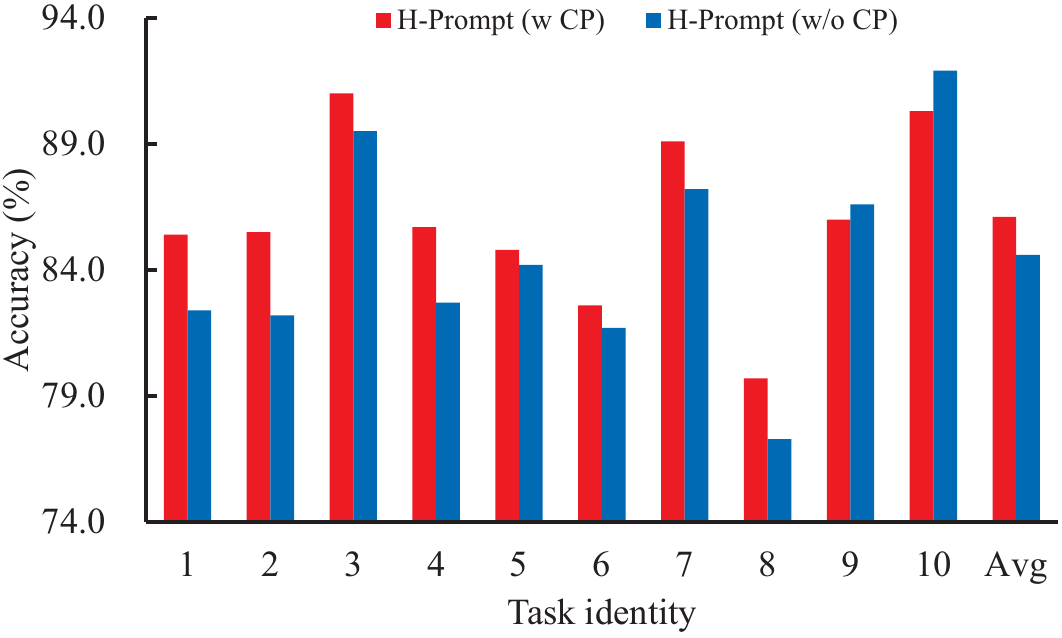}}
    \end{center}
    \caption{Comparison of the performance  about the model with the prompts of the $T$-th (final) task  between H-Prompts  with class prompt (w CP)  and H-Prompts  without class prompt (w/o CP)  on Split CIFAR-100 with one task-aware key per task.}
    \label{fig5}
    \end{figure}

\begin{figure*}[t]
  \centering
  \begin{subfigure}{0.2\linewidth}
    \includegraphics[width=1\linewidth]{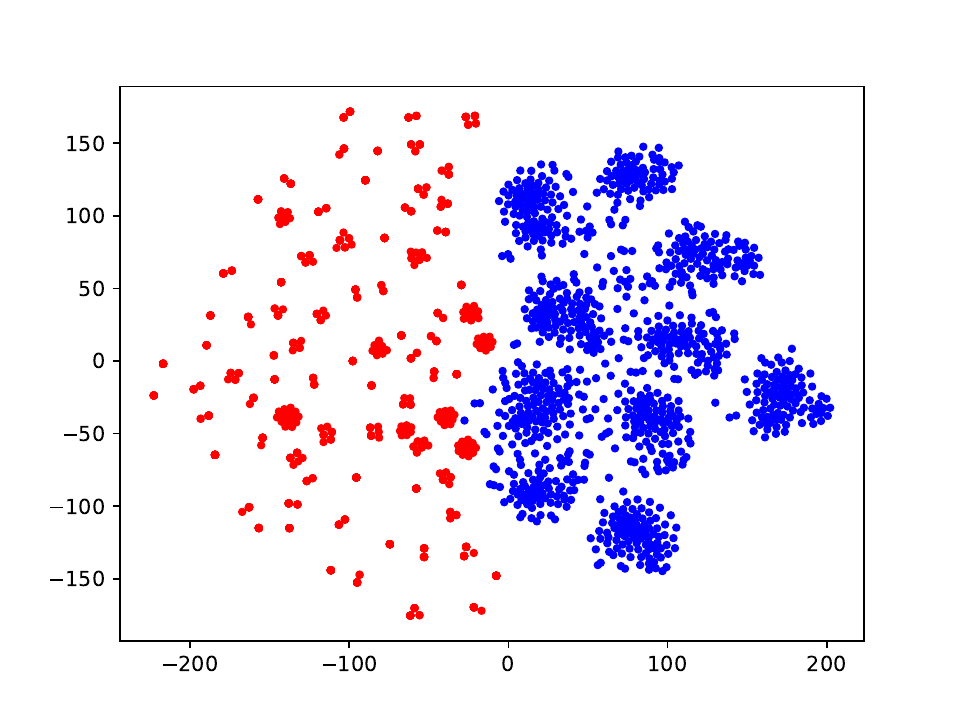}
    \caption{}
  \end{subfigure}
  \hspace{3mm}
  \begin{subfigure}{0.2\linewidth}
  \includegraphics[width=1\linewidth]{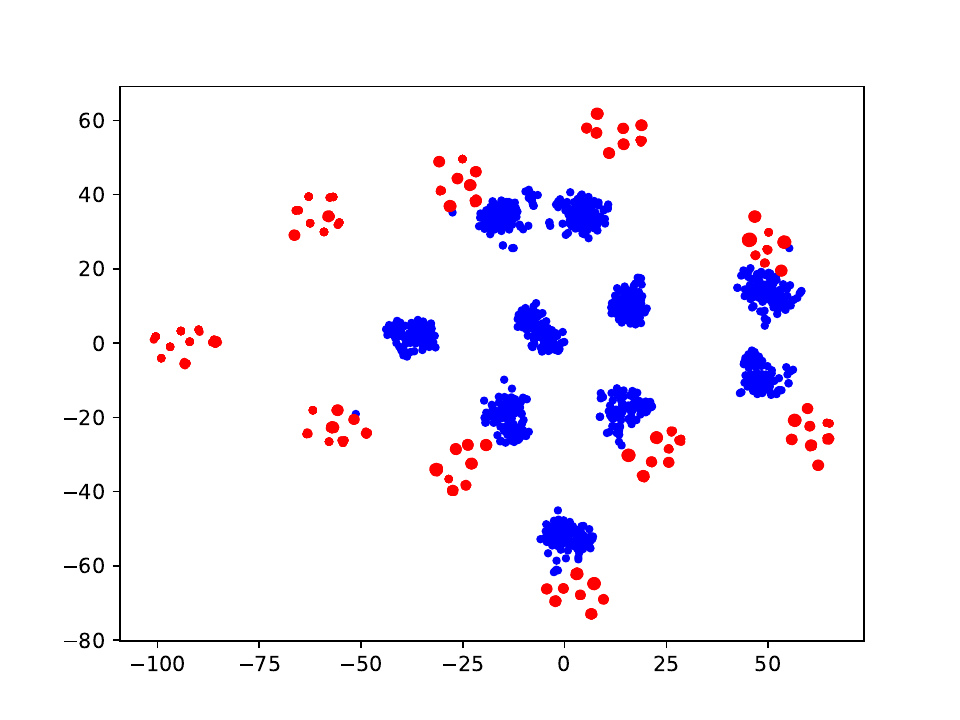}
  \caption{}
  \end{subfigure}
  \hspace{3mm}
  \begin{subfigure}{0.2\linewidth}
  \includegraphics[width=1\linewidth]{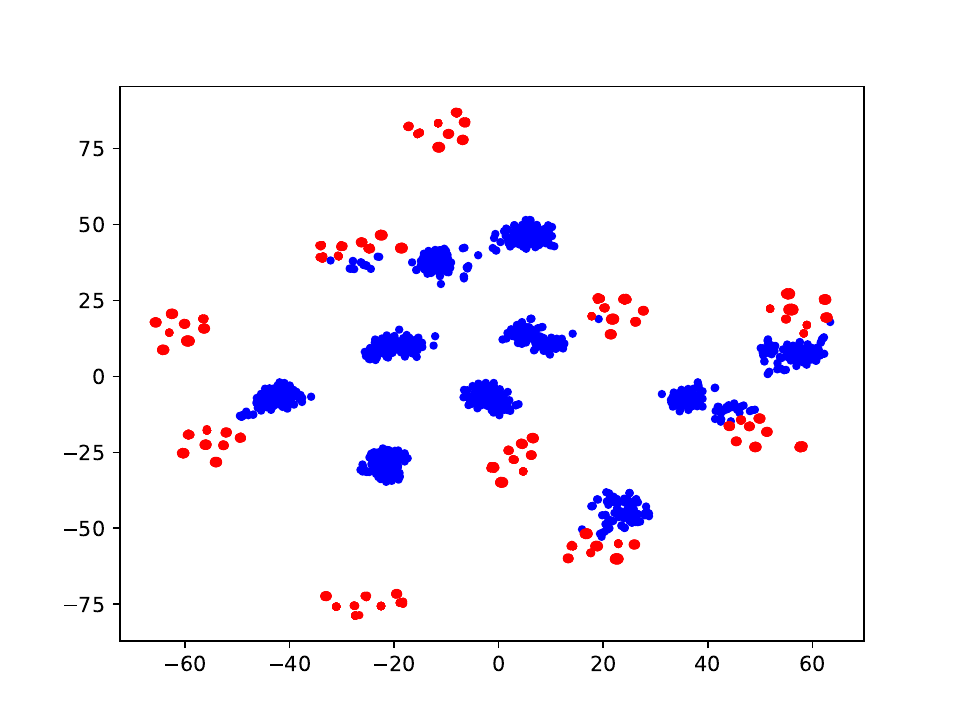}
  \caption{}
  \end{subfigure}
  \hspace{3mm}
  \begin{subfigure}{0.2\linewidth}
  \includegraphics[width=1\linewidth]{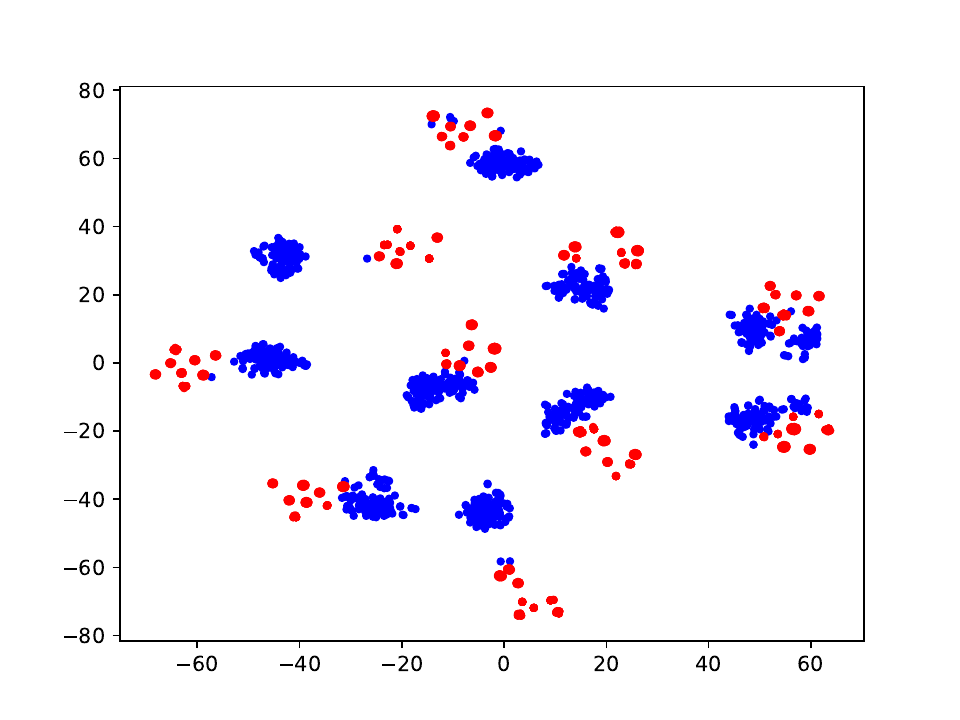}
  \caption{}
  \end{subfigure}
  \caption{Visualization of virtual representations replayed by class prompts and real representations of images   in the $1$-st task  on Split CIFAR-100.  (a), (b), (c), and (d) depict the change of representations along with the number of epoch increases.  Red and Blue dots represent the real  representations of real images and the virtual representations of class prompts, respectively.}
  \label{fig7}
  \end{figure*}

\textbf{Analysis of inference strategy.}
We analyze the effectiveness of the inference strategy (\emph{i.e.,} task-aware query-key mechanism) in H-Prompts on Split CIFAR-100, as shown in Table~\ref{tab4}.
The naive key-value mechanism, as utilized in Dualprompt~\cite{a21}, directly leverages the pre-trained backbone as a query model to obtain keys. 
In contrast, the task-aware query-key mechanism uses the pre-trained backbone integrated with task prompt and general prompt to yield task-specific keys. 
The table reveals that the task-aware query-key mechanism outperforms the naive one, exhibiting a 0.7\% higher average accuracy and a 0.3\% lower forgetting rate on Split CIFAR-100, thus demonstrating the advantage of the proposed mechanism.

\textbf{H-Prompts with shared general prompt.}
 To acquire generalized knowledge, the $i$-th  general prompt $\mathbf{g}_i$ is obtained by applying self-supervised learning of the data in the $i$-th task $\mathcal{T}_i$.
 In this section, we explore gaining a shared general prompt $\mathbf{g}$ with sequential tasks $\{\mathcal{T}_i\}_{i=1}^{T}$ in H-Prompts, \emph{i.e.}, utilizing the general prompt of the $T$-th (final) task for inference.
 The performances of H-Prompts and H-Prompts with a shared general prompt are contrasted in Table~\ref{H-Prompts (w. sgp)}. 
 Interestingly, both versions deliver similar results, such as a minor 0.1\% gap in average accuracy on Split CIFAR-100 and a 0.6\% gap on Split ImageNet-R. 
 When the generalized knowledge of the shared general prompt, $\mathbf{g}$, is applied across all tasks, it delivers consistent performance, underscoring the power of self-supervised learning for acquiring generalized knowledge. 
 Nevertheless, H-Prompts outperforms its variant with a shared general prompt on both Split CIFAR-100 and Split ImageNet-R, indicating that the generalized knowledge obtained from self-supervised learning  of the data in each task  promotes  the learning of  corresponding task.

 \begin{figure}[t]
  \centering

  \begin{subfigure}{0.47\linewidth}
      \includegraphics[width=1\linewidth]{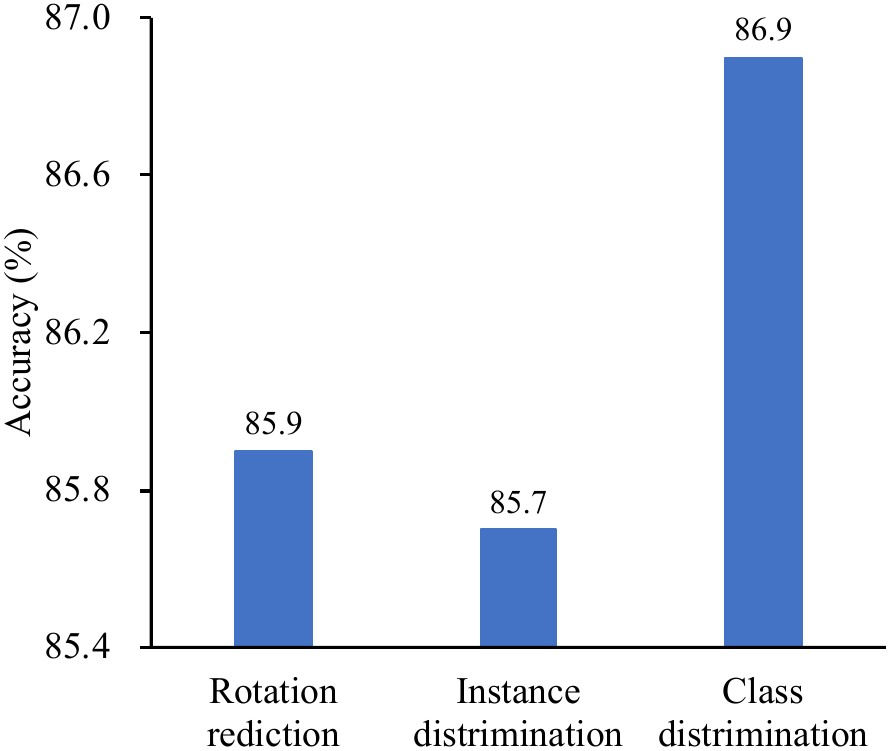}
      \caption{Average accuracy}
    \end{subfigure}
  \begin{subfigure}{0.47\linewidth}
    \includegraphics[width=1\linewidth]{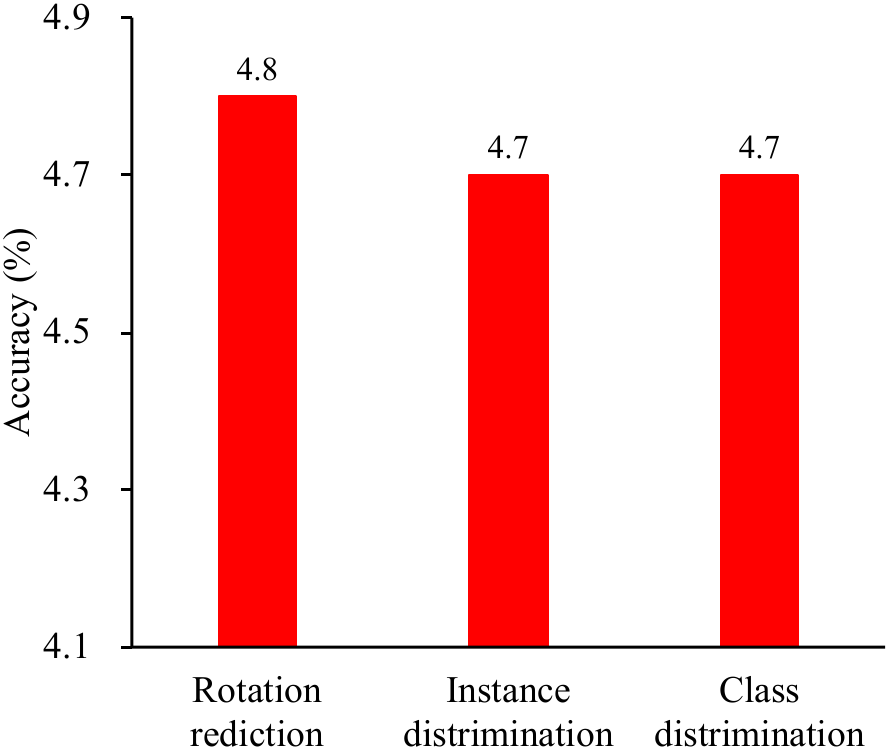}
    \caption{Forgetting}
  \end{subfigure}

  \caption{Analyses of different pretext tasks for self-supervised learning on Split CIFAR-100 with one key per task.}
  \label{fig3}
\end{figure}

\textbf{Analysis of class prompt for the final  prompts.}
To ascertain the impact of the class prompt in preserving past knowledge, we conduct an evaluation utilizing the prompts of the $T$-th (final) task  with  and without considering class prompt, as illustrated in Figure~\ref{fig5}. 
When employing the class prompt, H-Prompts (w CP) consistently outperforms H-Prompts (w/o CP) on most preceding task datasets, demonstrating that the class prompt effectively mitigates the forgetting of past task knowledge. 
Notably, as H-Prompts (w/o CP) solely concentrates on current task knowledge, its performances in later tasks (Tasks 9 and 10) exceed that of H-Prompts (w CP). 
Additionally, the overall average performance of H-Prompts (w CP) at 86.1\% surpasses the 84.6\% performance of H-Prompts (w/o CP), further underscoring the necessity of the class prompt.

\textbf{Visualization of class prompt  representation.}
We visualize the virtual representations replayed by  class prompts and real representations of images  from all classes in the $1$-st task, as shown in  Figure~\ref{fig7}. 
Over time, the class prompt representations gradually align with the real image representations. 
Furthermore, the virtual representations of different classes ultimately separate distinctively.  
These results confirm that the class prompt accurately models the class distribution in each task, thus effectively replaying past task knowledge to avoid catastrophic forgetting.

\begin{figure}[t]
  \centering
  
  \begin{subfigure}{0.48\linewidth}
    \includegraphics[width=1\linewidth]{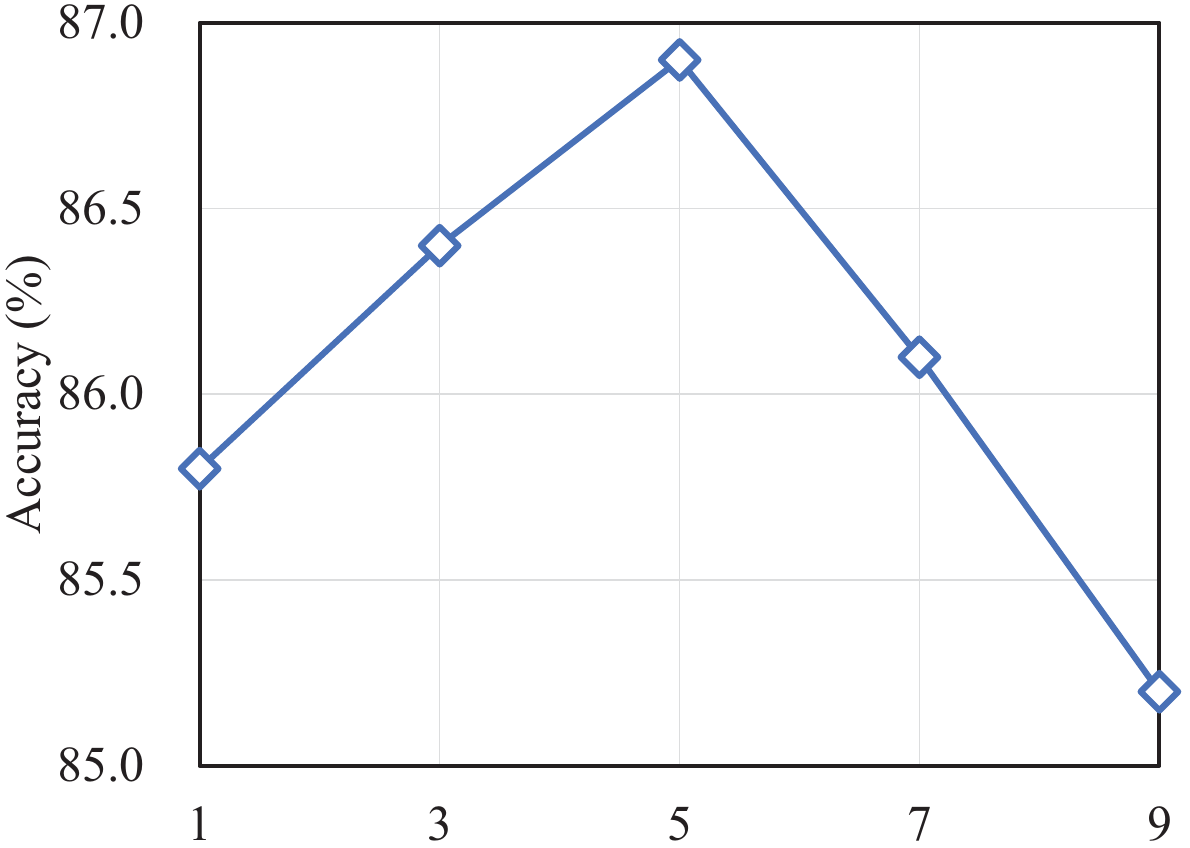}
    \caption{\footnotesize{The length of task prompt $L_t$}}
  \end{subfigure}
  \begin{subfigure}{0.47\linewidth}
  \includegraphics[width=1\linewidth]{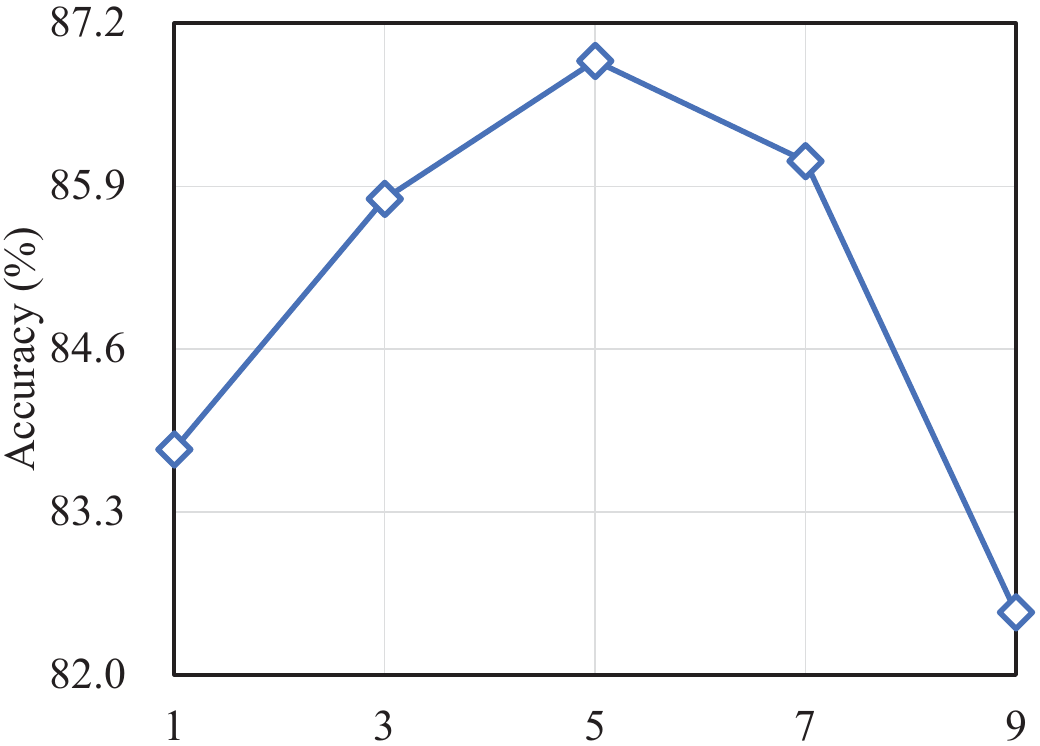}
  \caption{\footnotesize{The depth of task prompt $\varGamma_t$}}
  \end{subfigure}
  \caption{Hyper-parameter analyses of the length of task prompt $L_t$ and the depth of task prompt $\varGamma_t$ on Split CIFAR-100 with one task-aware key per task.}
  \label{fig6-t}
  \end{figure}

  \textbf{Effect of different pretext task.}
  In this paper, we employ class discrimination as the pretext task for self-supervised training, the efficacy of which is demonstrated in the ablation study of H-Prompts. 
  For the sake of experimental comprehensiveness, we additionally implement rotation prediction~\cite{a30} and instance discrimination~\cite{a28} as pretext tasks for self-supervised training. 
  The  results of three different pretext task with one key per task are presented in Figure~\ref{fig3}. 
Both rotation prediction and instance distrimination exhibit lower average accuracies than class discrimination, dropping by 1.0\% and 1.2\% respectively, with similar forgetting rates.
  Rotation prediction  implements  image rotation and  predicts the rotation angle for each image.
  Nevertheless, the optimal representation for rotation predictions is incapable of improving the discrimination in each class.
  Moreover, instance distrimination simply separates each image ignoring class infomation, so  the images in the same class are also pushed apart, resulting in less discriminative representation. 
  Conversely,  class distrimination  utilizes class information for self-supervised training to learn class-invariant representation, promoting generalized discriminative representation learning.
  The above results emphasize that  the pretext task of the self-supervised learning requires careful selection to acquire generalized features for incremental learning.

\textbf{Hyper-parameter analyses of task prompt.}
We perform hyper-parameter analyses about the length of task prompt $L_t$ and the depth of task prompt $\varGamma_t$ on Split CIFAR-100, as illustrated in Figure~\ref{fig6-t}. 
The results show that the model performs optimally when $L_t$ = 5 and $\varGamma_t$ = 5, indicating the importance of choosing appropriate $L_t$ and $\varGamma_t$ for optimal results. Larger task prompt lengths $L_t$ and task prompt depths $\varGamma_t$ lead to an excessive number of parameters, which may cause overfitting to the current task. Conversely, smaller $L_t$ and $\varGamma_t$ may result in inadequate model plasticity, thus hindering the learning of new task knowledge.

  \begin{figure}
    \centering
    
    \begin{subfigure}{0.48\linewidth}
      \includegraphics[width=1\linewidth]{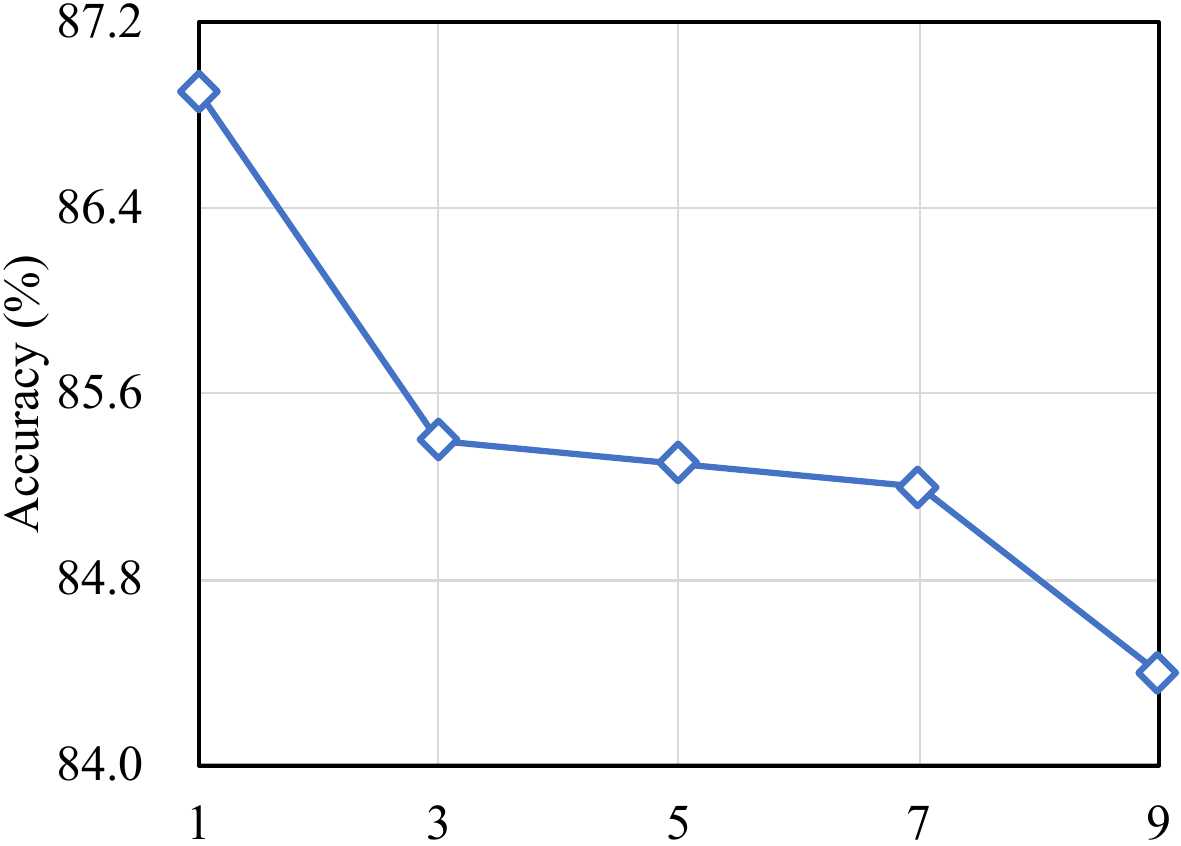}
      \caption{\footnotesize{The length of task prompt $L_g$}}
    \end{subfigure}
    \begin{subfigure}{0.47\linewidth}
    \includegraphics[width=1\linewidth]{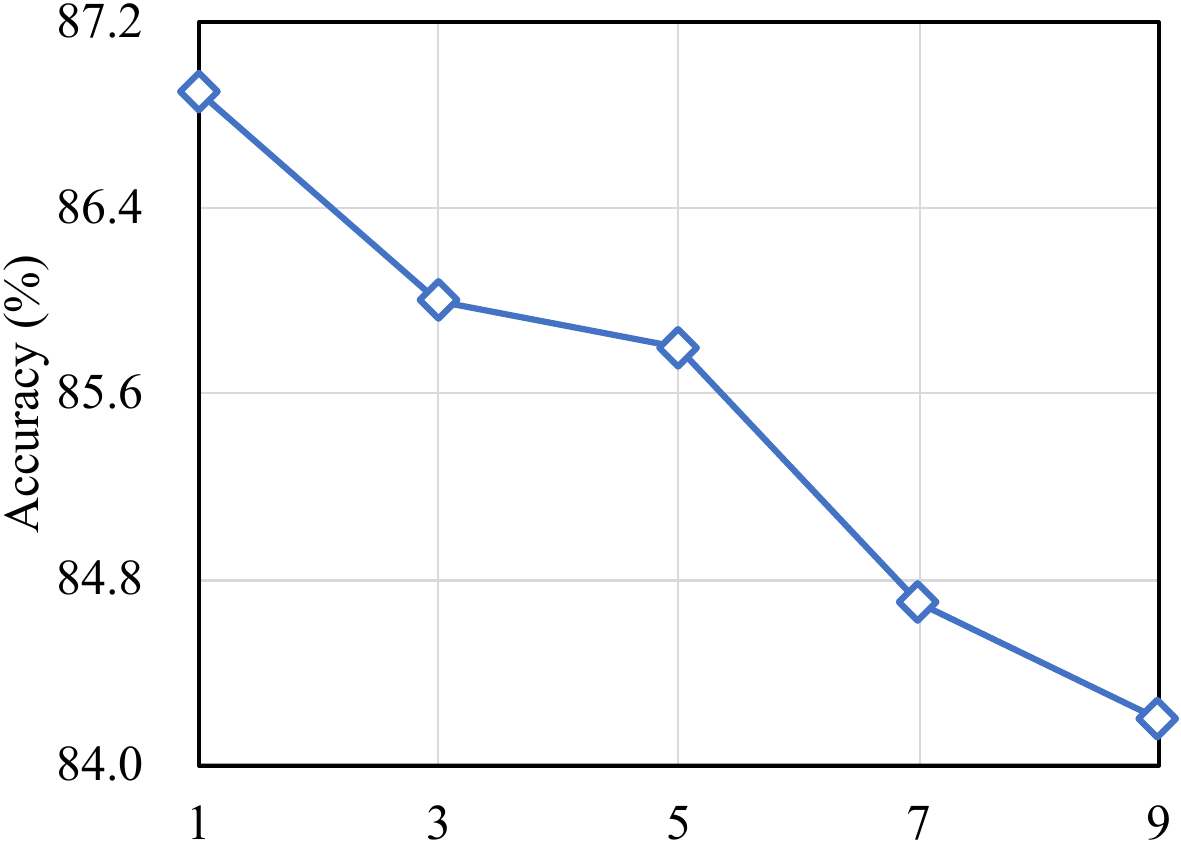}
    \caption{\footnotesize{The depth of task prompt $\varGamma_g$}}
    \end{subfigure}
    \caption{Hyper-parameter analyses of the length of general prompt $L_g$ and the depth of general prompt $\varGamma_g$ on Split CIFAR-100 with one task-aware key per task.}
    \label{fig6-g}
    \end{figure}

\textbf{Hyper-parameter analyses of general prompt.}
We also conduct hyper-parameter analyses about the length of general prompt $L_g$ and the depth of general prompt $\varGamma_g$ on Split CIFAR-100, as illustrated in Figure~\ref{fig6-g}. 
Observations indicate that the model performs optimally when both $L_g$ and $\varGamma_g$ are set to 1. 
Nonetheless, the model's performance declines when either $L_g$ or $\varGamma_g$ is increased. 
For instance, an increase of the length of the general prompt $L_g$ to 3 leads to a 1.5\% performance drop, and increasing the depth of the general prompt $\varGamma_g$ to 3 results in a 0.5\% performance decline. 
Furthermore, the model's performance continually falls as the length of general prompt $L_g$ or the depth of general prompt $\varGamma_g$ increases.
These results imply that although highly generalized knowledge acquired through self-supervised learning can help alleviate catastrophic forgetting in continual learning, an excessive amount of generalized knowledge can interfere with the model's ability to learn specific knowledge for downstream tasks.

\begin{table}[t]
  \centering
  \scriptsize
  \setlength{\tabcolsep}{8pt}
  \footnotesize
  \renewcommand{\arraystretch}{1}
  \caption{Domain incremental learning on CDDB-Hard.}
  \label{tab6}
  
  {
  \begin{tabular}{l|| c| c | c c  }
    \shline
    \multirow{1}{*}{Methods} &\multirow{1}{*}{Buffer size}& \multirow{1}{*}{Average Acc $\uparrow$}& \multirow{1}{*}{Forgetting $\downarrow$}\\
     \hline

    EWC~\cite{a8}&  0& 50.6 &  42.6  \\
    LwF~\cite{a60}&  0& 60.9 & 13.5   \\
    Dytox~\cite{a43}&  0& 51.3 & 45.9   \\
    L2P~\cite{a20}&  0& 61.3 & 9.2   \\
    S-iPrompts~\cite{a72} &  0& 74.5 & 1.3   \\

    \textbf{H-Prompts} &  0& \textbf{76.1} & \textbf{0.9}   \\
    \hline
    Upper-bound &  -& 85.5 & -  \\

  \shline
  \end{tabular}}
\end{table}

\begin{figure}[t]
  \centering

  \begin{subfigure}{0.47\linewidth}
      \includegraphics[width=1\linewidth]{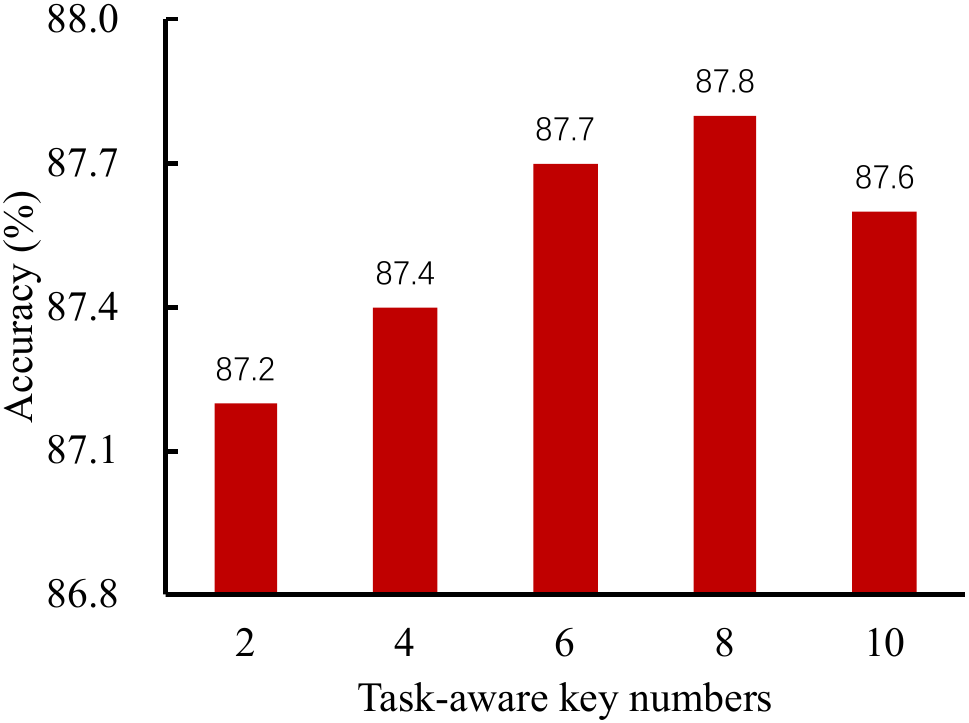}
      \caption{Split CIFAR-100}
    \end{subfigure}
  \begin{subfigure}{0.47\linewidth}
    \includegraphics[width=1\linewidth]{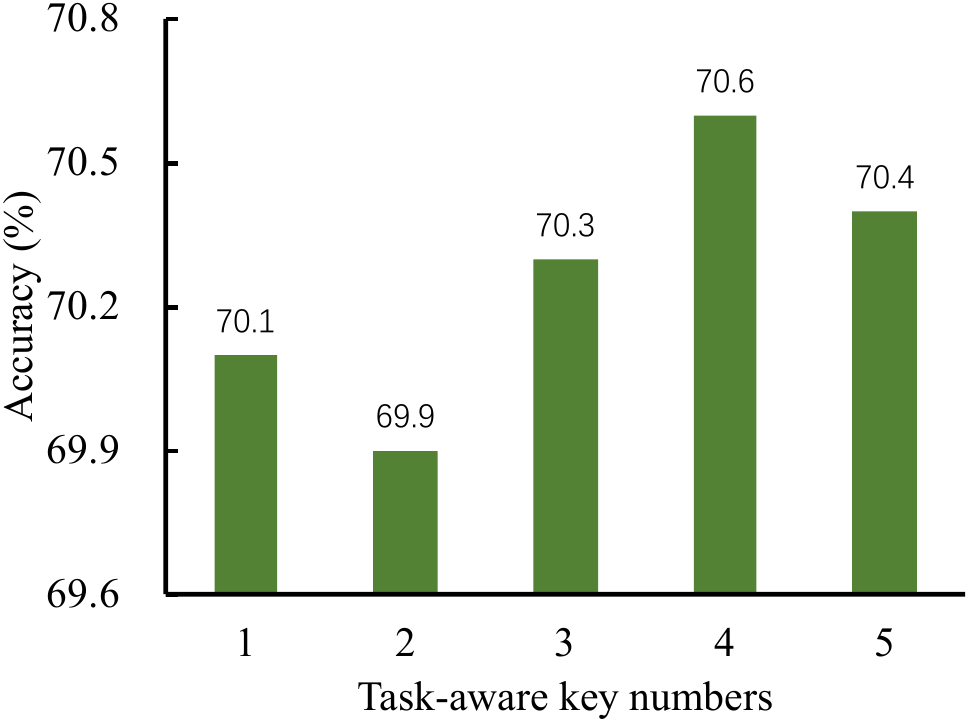}
    \caption{Split ImageNet-R}
  \end{subfigure}

  \caption{Analyses of different task-aware key numbers on Split CIFAR-100 and Split ImageNet-R.}
  \label{key-numbers}
\end{figure}

\textbf{Hyper-parameter analyses of task-aware key numbers.}
We conduct hyper-parameter analyses concerning the number of task-aware keys $o$ on Split CIFAR-100 and Split ImageNet-R, as shown in Figure~\ref{key-numbers}. 
The results demonstrate that the model performs best when the number of task-aware keys $o$ is set to 8 per class and 4 per class for Split CIFAR-100 and Split ImageNet-R, yielding accuracies of 87.8\% and 70.6\%, respectively. 
However, the model's performance declines when $o$ deviates from this optimal value. This finding underscores the significance of selecting an appropriate number of task-aware keys $o$. 
If $o$ is too low, the keys obtained may fail to adequately capture the task's data distribution, leading to insufficient task knowledge representation. 
Conversely, a high $o$ may lead to overfitting to the task knowledge in the training data, as it may capture rare data points in the data distribution, thus reducing generalization to the test data.

\textbf{Model generalization.}
Targeting to illustrate the generalization of the proposed H-Prompts, we implement experiments about domain incremental learning setting on CDDB-Hard~\cite{a73}, as shown in Table~\ref{tab6}. 
We observe that H-Prompts achieves the best results on both average accuracy and forgetting evaluation metrics, outperforming the state-of-the-art domain incremental learning method S-iPrompts~\cite{a72} by 1.6\% of average accuracy and 0.4 of forgetting.
The superior performance of H-Prompts demonstrates the generalization and effectiveness of H-Prompts.

\section{Conclusion}
In this paper, we propose a novel Hierarchical Prompts (H-Prompts) rehearsal-free paradigm to overcome the catastrophic forgetting of prompts for continual learning.
H-Prompts  comprises class prompt, task prompt, and general prompt, which model the distribution of classes in each task, capture past task knowledge and current task knowledge, and learn generalized  knowledge, respectively.
Evaluations reveal that H-Prompts outperforms other methods on two standard benchmarks of class incremental learning, attesting to the effectiveness of the proposed H-Prompts. 
Future work will explore the potential of applying H-Prompts to real-world applications, such as object detection and semantic segmentation.


%





\ifCLASSOPTIONcaptionsoff
  \newpage
\fi



\bibliographystyle{IEEEtran}
\bibliography{IEEEabrv,egbib}

\end{document}